\documentclass[10pt,twocolumn,letterpaper]{article}

\usepackage{cvpr}              

\usepackage[dvipsnames]{xcolor}

\usepackage{graphicx}
\usepackage{amsmath}
\usepackage{amssymb}
\usepackage{booktabs}

\usepackage{times}
\usepackage{epsfig}
\usepackage{bbm}

\usepackage{verbatim}
\usepackage{listings}

\usepackage{algorithm}
\usepackage{algorithmicx}
\usepackage{algpseudocode}

\usepackage{color}

\usepackage{ctable}
\usepackage{makecell}
\usepackage{tabularx}
\usepackage{multirow}
\usepackage{multicol}
\usepackage{arydshln}

\usepackage{pifont}
\newcommand{\cmark}{\ding{51}}%
\newcommand{\xmark}{\ding{55}}%

\newcommand{\ourmethod}{TeHOR}

\definecolor{cvprblue}{rgb}{0.21,0.49,0.74}
\usepackage[pagebackref,breaklinks,colorlinks,allcolors=cvprblue]{hyperref}

\def\paperID{*****} 
\def\confName{CVPR}
\def\confYear{2026}

\title{\ourmethod: Text-Guided 3D Human and Object Reconstruction with Textures}

\author{
  Hyeongjin Nam$^{1}$ \hskip1.6em Daniel Sungho Jung$^{2}$ \hskip1.6em Kyoung Mu Lee$^{1,2}$ \vspace{+2mm} \\ 
   $^{1}$Dept. of ECE\&ASRI, $^{2}$IPAI, Seoul National University  \\
   {\tt\small \{namhjsnu28, dqj5182, kyoungmu\}@snu.ac.kr} \\
   \small{\url{https://hygenie1228.github.io/TeHOR/}}
}

\begin{document}
\twocolumn[{
\maketitle
{\centering
\includegraphics[width=1.0\linewidth]{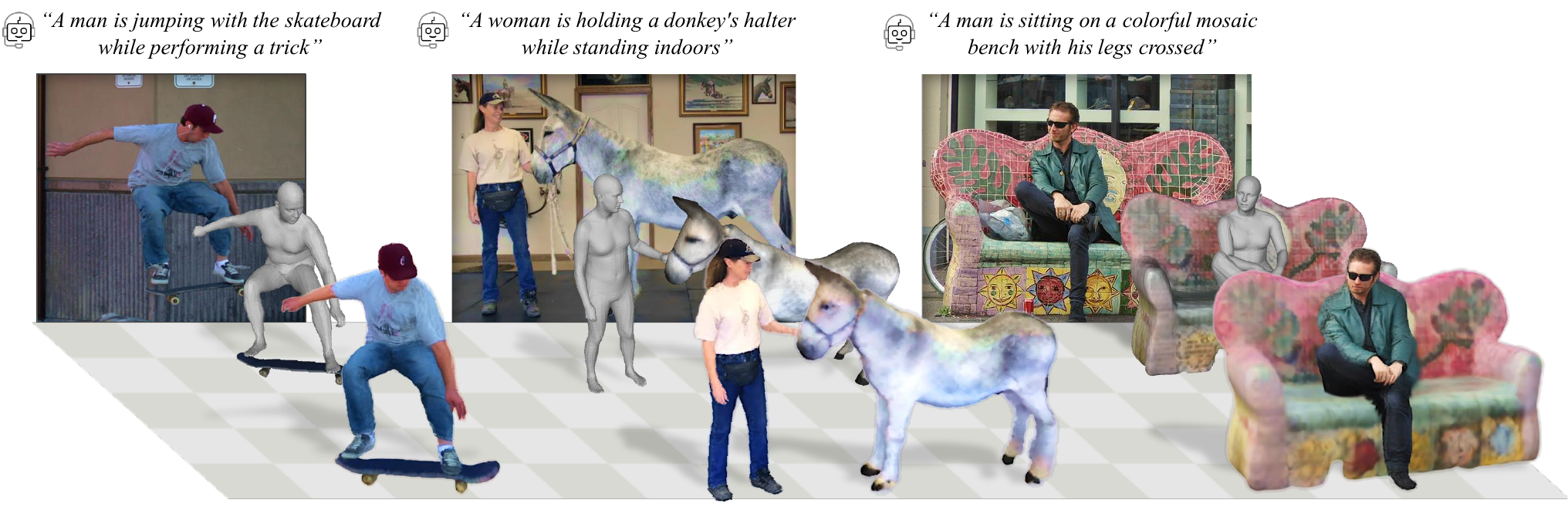}}
\vspace{-1.5em}
\captionsetup{hypcap=false}
\captionof{figure}
{
    \textbf{\ourmethod.}
    Given a single image, our framework jointly reconstructs textured 3D human and object by capturing their holistic and semantic interactions using text descriptions.
}
\label{fig:0_teaser}
\vspace{+2.5em}
}]

\begin{abstract}
Joint reconstruction of 3D human and object from a single image is an active research area, with pivotal applications in robotics and digital content creation.
Despite recent advances, existing approaches suffer from two fundamental limitations.
First, their reconstructions rely heavily on physical contact information, which inherently cannot capture non-contact human–object interactions, such as gazing at or pointing toward an object.
Second, the reconstruction process is primarily driven by local geometric proximity, neglecting the human and object appearances that provide global context crucial for understanding holistic interactions.
To address these issues, we introduce \textbf{\ourmethod}, a framework built upon two core designs.
First, beyond contact information, our framework leverages text descriptions of human–object interactions to enforce semantic alignment between the 3D reconstruction and its textual cues, enabling reasoning over a wider spectrum of interactions, including non-contact cases.
Second, we incorporate appearance cues of the 3D human and object into the alignment process to capture holistic contextual information, thereby ensuring visually plausible reconstructions.
As a result, our framework produces accurate and semantically coherent reconstructions, achieving state-of-the-art performance.
\end{abstract}
\section{Introduction}
\label{sec:introduction}
\begin{figure}[t!]
  \centering
  \includegraphics[width=1.0\linewidth]{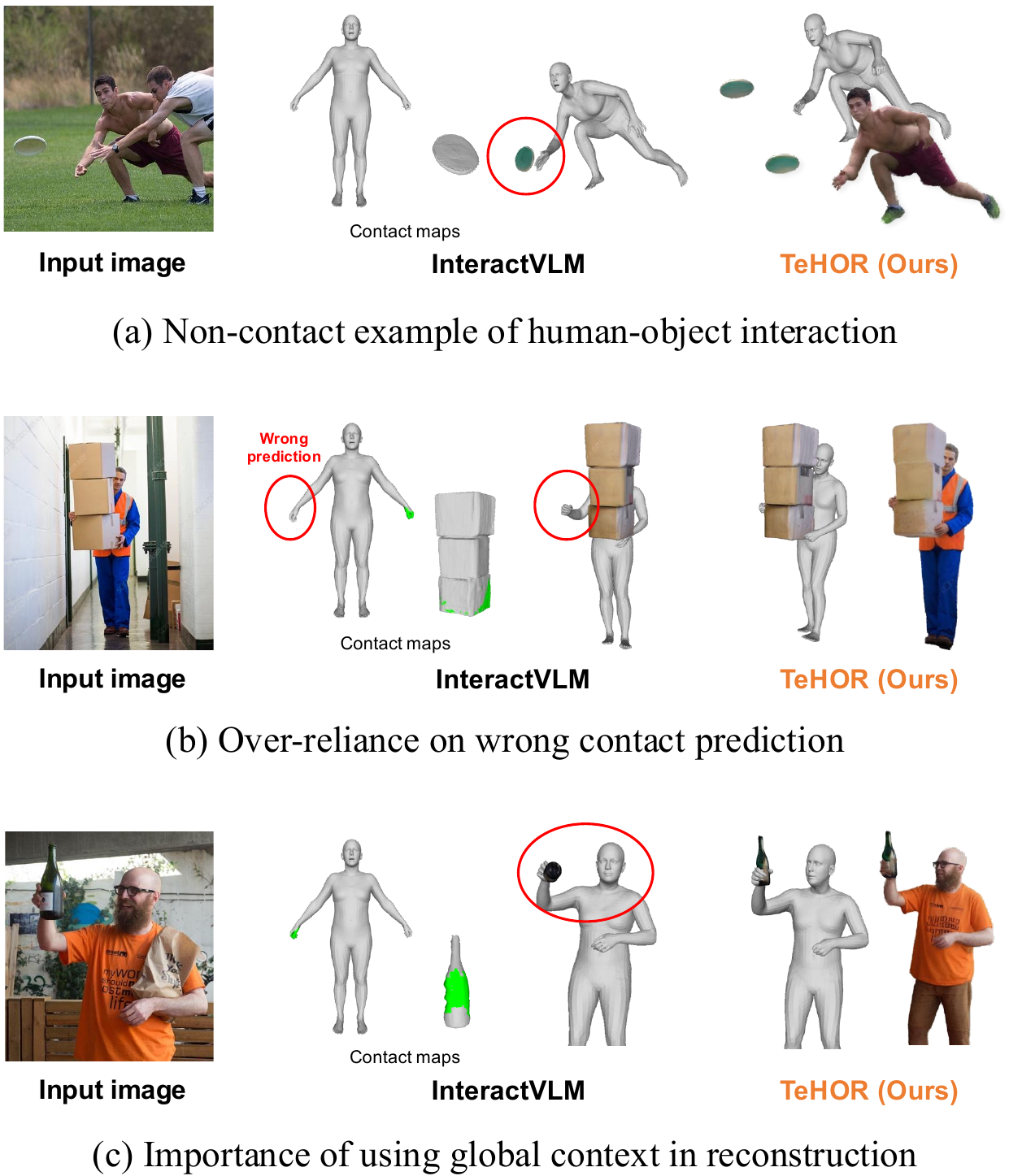}
  \vspace*{-1.3em}
  \caption{\textbf{Limitations of existing reconstruction methods.}
  Previous methods suffer from over-reliance on contact information and neglect the global interaction context, leading to implausible reconstructions.
  In contrast, \ourmethod~produces accurate and plausible 3D reconstructions by leveraging holistic and semantic guidance from text descriptions.
  }
  \vspace*{-0.0em}
  \label{fig:1_introduction}
\end{figure}
Joint reconstruction of 3D human and object from a single image is a cornerstone of human behavior understanding, as it provides insight into how humans interact physically and semantically with their surroundings, enabling broad applications in robotics, AR/VR, and digital content creation.
To achieve physically and semantically accurate reconstruction, it is essential to faithfully capture human–object interaction (HOI), ranging from explicit physical contacts, such as grasping a cup, to implicit non-contact relations, such as reaching toward a door handle or gazing at a monitor.

Most existing methods for 3D human and object reconstruction from a single image~\cite{xu2021d3d,xie2022chore,nam2024contho,xie2023templatefree,dwivedi2025interactvlm,cseke2025pico,wang2025end,li2025scorehoi} have primarily utilized human–object contact information as a major cue for interaction reasoning.
These approaches first predict contact regions on the 3D human and object surfaces, and subsequently enforce local geometric proximity at the contact regions through iterative fitting.
While these methods have shown notable progress, they suffer from two fundamental limitations.
First, previous methods rely heavily on contact information, which provides limited physical cues and cannot capture non-contact interactions common in the real world, such as gazing at or pointing toward an object.
As illustrated in~\cref{fig:1_introduction}~(a), when a person prepares to catch a frisbee, the absence of physical contact provides no cues for the reconstruction system to leverage, resulting in incorrect 3D reconstructions.
Even in contact scenarios, over-reliance on contact information also makes these methods vulnerable to contact prediction errors, as shown in~\cref{fig:1_introduction}~(b), where inaccurate contact prediction directly leads to erroneous reconstruction results.
Second, the fitting processes of the existing methods are primarily driven by local geometric proximity, ignoring global appearance cues of the human and object.
The appearance cues, such as color and shading, provide rich contextual information for understanding the holistic structure of human–object interactions.
However, the previous reconstruction systems primarily fit the geometric proximity between human and object without considering the holistic appearance, resulting in globally implausible results, such as an incorrectly oriented bottle and a misaligned human gaze in \cref{fig:1_introduction}~(c).

To address these limitations, we propose \textbf{\ourmethod}~(\textbf{Te}xt-guided 3D \textbf{H}uman and \textbf{O}bject \textbf{R}econstruction), which leverages rich text descriptions of human-object interaction as strong guidance for reconstruction.
Given a single image, our framework first extracts text descriptions that specify the human–object interaction depicted in the image, using a vision–language model (\textit{e.g.}, GPT-4~\cite{achiam2023gpt}).
Based on the text descriptions, we jointly optimize the geometry and texture of the 3D human and object by supervising their rendered 2D appearances to align them with the semantic cues in the text.
Specifically, this supervision is implemented by utilizing a pre-trained diffusion network (\textit{e.g.}, StableDiffusion~\cite{rombach2022high}) conditioned on the text descriptions.
As the diffusion network encompasses strong prior knowledge of the association between visual appearance and text descriptions, it serves as a bridge between textual and visual domains.
During the optimization, the diffusion network computes text-conditioned score gradients that drive the rendered appearances toward the visual distribution conditioned on the text, progressively refining the 3D human and object to better reflect the described interaction.
This optimization is performed across multiple viewpoints, encouraging the reconstructed 3D human and object to exhibit consistent semantics across different viewpoints.

Built upon this design, our text-guided optimization offers two key advantages.
First, as text descriptions contain semantic cues that extend beyond human-object contact, our framework enables reasoning about a wide range of interactions, including non-contact scenarios, such as approaching to catch a frisbee.
Second, by supervising the 2D appearances of the human and object with a pre-trained diffusion network, our framework takes into account the holistic visual plausibility of the interaction, unlike previous methods that rely primarily on local geometric proximity.
As a result, we show that our framework significantly improves the accuracy and plausibility of 3D human and object reconstruction and achieves state-of-the-art performance in both general and non-contact scenarios.
Moreover, to the best of our knowledge, our framework is the first to jointly reconstruct full 3D textures of the human and the interacting object, which enables the creation of immersive and realistic 3D digital assets.

Our contributions can be summarized as follows.
\begin{itemize}
\item We propose \textbf{\ourmethod}, which jointly reconstructs 3D human and object by leveraging text descriptions as semantic guidance, extending beyond physical contact cues.
\item To capture the holistic context of human–object interaction, we incorporate appearance cues from the human and object by globally aligning their rendered appearances with the textual descriptions.
\item Extensive experiments demonstrate that our proposed framework significantly outperforms previous reconstruction methods across diverse interaction scenarios.
\end{itemize}

\section{Related works}
\label{sec:related_works}
\noindent\textbf{3D human and object reconstruction.}
Most of the recent works on 3D human and object reconstruction~\cite{chen2019holistic++,zhang2020perceiving,xie2024rhobin,xu2021d3d,xie2022chore,nam2024contho,gavryushin2024romeo,xie2023templatefree,li2025scorehoi} primarily focus on capturing local interaction priors, particularly contact, to refine the spatial relationship between 3D human and object.
PHOSA~\cite{zhang2020perceiving} optimizes the 3D spatial arrangement of humans and objects based on pre-defined contact labels, encouraging the contact regions on both human and object to ensure geometric proximity.
CONTHO~\cite{nam2024contho} is a framework that predicts contact maps directly from images and leverages a contact-based Transformer to enhance reconstruction using the predicted contact maps.
More recently, several methods have been proposed for open-vocabulary reconstruction, enabling strong scalability and generalization to object categories unseen during training.
InteractVLM~\cite{dwivedi2025interactvlm} infers 2D contact maps in multiple views through a fine-tuned vision–language model~\cite{liu2023llava} and uses the contact information to guide reconstruction.
HOI-Gaussian~\cite{wen2025reconstructing} integrates a contact loss with an ordinal depth loss to more accurately capture relative depth between humans and objects.

Despite these advances, existing methods still rely heavily on physical cues, such as contact, causing them to struggle with complex interactions that require semantic reasoning.
InteractVLM depends on accurate contact estimation, and failures in contact prediction directly degrade the quality of the final 3D reconstruction.
While HOI-Gaussian introduces an ordinal depth constraint to mitigate contact failure, it remains vulnerable to complex interaction scenarios, such as severe human–object occlusion, as it relies on front-view cues without semantic reasoning.
Unlike these methods, our framework is capable of holistic semantic reasoning about human-object interactions beyond contact information by leveraging global contextual information from text descriptions.
Additionally, our framework goes beyond prior methods by reconstructing textured 3D humans and objects that provide richer representations than the non-textured surfaces, benefiting immersive downstream applications such as AR/VR.

\noindent\textbf{Human-object interaction.}
With the emergence of large-scale 3D datasets~\cite{bhatnagar2022behave,huang2022intercap,wen2025reconstructing}, numerous studies~\cite{hassan2021populating,huang2022capturing,tripathi2023deco,yang2024lemon,shimada2022hulc,han2023chorus,chen2023detecting,kim2025david} have focused on capturing human–object interactions.
Human–object contact has been a common cue in this line of research due to its intuitive physical representation observable in images.
BSTRO~\cite{huang2022capturing} is a Transformer-based~\cite{devlin2019bert} framework that predicts dense body contact from a single image.
DECO~\cite{tripathi2023deco} utilizes a cross-attention network that jointly leverages human body parts and 2D scene context for contact estimation.
LEMON~\cite{yang2024lemon} extends contact estimation approaches by modeling relations among human contact, object affordance, and spatial configuration.
More recently, ComA~\cite{kim2024beyond} introduces a probabilistic affordance representation that extends binary contact to incorporate relative orientation and proximity.
Our framework goes beyond this line of work by leveraging textual descriptions that capture the semantic context of human-object interaction, enabling plausible 3D reconstructions.

\noindent\textbf{3D human reconstruction.}
Most of the 3D human reconstruction methods are built upon parametric human models~(\textit{e.g.}, SMPL~\cite{loper2015smpl} and SMPL-X~\cite{pavlakos2019expressive}) to estimate human body~\cite{baradel2024multi,dwivedi2024tokenhmr,wang2025prompthmr,nam2023cyclic,patel2024camerahmr,kanazawa2018end,kocabas2021pare,li2022cliff,zhang2021pymaf,feng2021collaborative} or clothed humans~\cite{saito2020pifuhd,zheng2020pamir,alldieck2022phorhum,liao2023high,pan2025humansplat,xiu2023econ,qiu2025anigs,nam2025parte,sim2025persona,pan2024humansplat,moon20223d}.
ARCH~\cite{he2020arch,he2021arch++} reconstructs animatable clothed humans using an implicit function that encodes occupancy, normals, and colors for detailed geometry and appearance.
TeCH~\cite{huang2024tech} employs a text-to-image diffusion model~\cite{ruiz2023dreambooth} to match the reconstructed appearance of the human with the textual description of the input image.
LHM~\cite{qiu2025lhm} represents human texture via 3D Gaussians in canonical space, enabling high-quality reconstruction and pose-controlled animation.
In our framework, we utilize LHM for the initial 3D human reconstruction due to its strong generalization to in-the-wild scenarios.

\begin{figure*}[t!]
  \centering
  \includegraphics[width=1.0\linewidth]{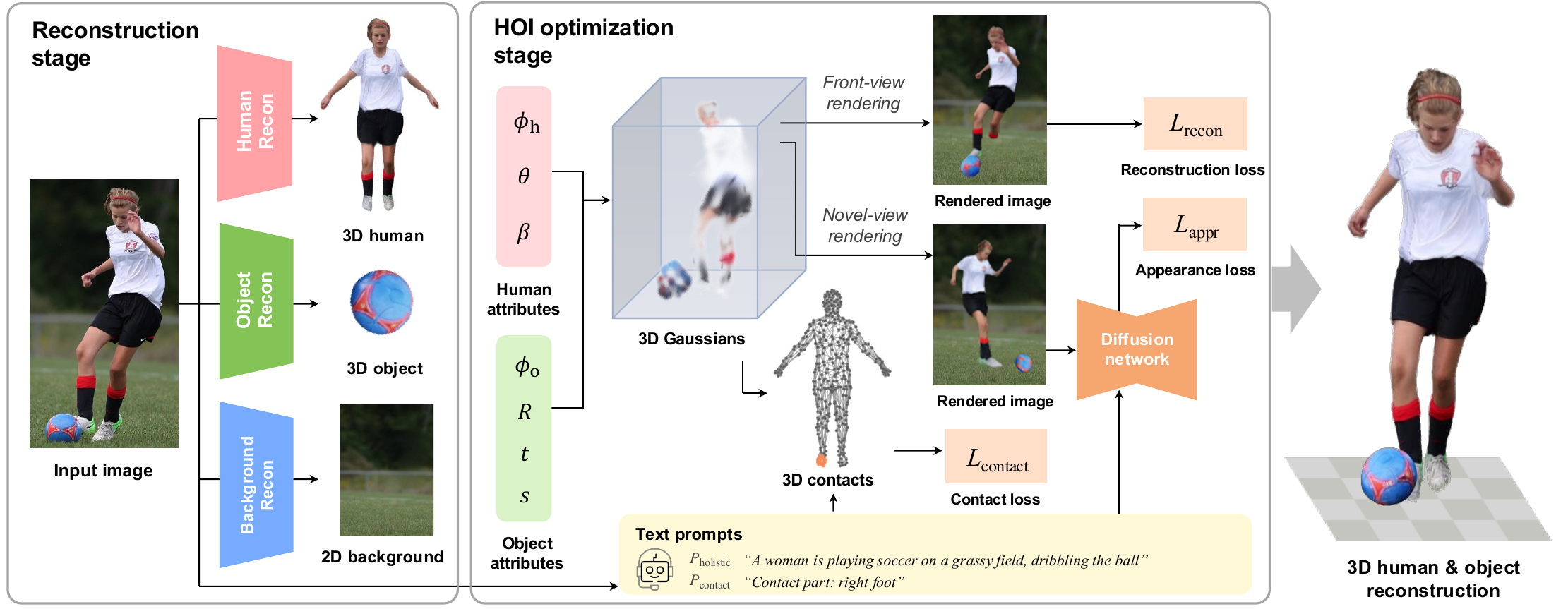}
  \vspace*{-1.5em}
  \caption{\textbf{Overall pipeline of \ourmethod.}
  Given an input image, our framework initially reconstructs a 3D human, a 3D object, and a 2D background. 
  Then, the initially reconstructed 3D human and object are jointly optimized using three core loss functions: reconstruction loss, appearance loss, and contact loss, to ensure accurate and semantically plausible human-object interaction.
  }
   \vspace*{+0.0em}
  \label{fig:2_pipeline}
\end{figure*}

\noindent\textbf{3D object reconstruction.}
Recent works on 3D object reconstruction~\cite{tang2023dreamgaussian,xu2024instantmesh,zhao2023michelangelo,liu2023zero,qian2023magic123,huang2024zeroshape,liu2023one2345,gao2024cat3d,long2024wonder3d,tang2024lgm,xiang2024structured} have increasingly focused on open-vocabulary settings to generalize beyond fixed object categories.
Zero-1-to-3~\cite{liu2023zero} introduces a viewpoint-conditioned image generative model that learns novel-view image synthesis, enabling zero-shot 3D reconstruction.
DreamGaussian~\cite{tang2023dreamgaussian} optimizes a 3D Gaussian representation~\cite{kerbl3Dgaussians} for efficient and high-quality reconstruction.
InstantMesh~\cite{xu2024instantmesh} is a feed-forward Transformer~\cite{vaswani2017attention} that exploits a multi-view diffusion prior for efficient textured 3D object reconstruction.
In all our experiments, we use the 3D object meshes produced by InstantMesh as the initial object reconstruction to ensure a fair and consistent baseline, given that most previous reconstruction methods are mesh-based pipelines.

\section{\ourmethod}
\label{sec:3_proposed_method}
\cref{fig:2_pipeline} illustrates the overall pipeline of \ourmethod.
In the following sections, we first describe the 3D representation of the human and object (\cref{sec:3.1_representation}).
Subsequently, we provide detailed descriptions of two stages: the reconstruction stage (\cref{sec:3.2_reconstruction_stage}) and the HOI optimization stage (\cref{sec:3.3_hoi_optimization}).

\subsection{3D representation}
\label{sec:3.1_representation}
We represent the 3D human and object, each as a set of 3D Gaussians, denoted by $\Phi_{\text{h}}$ and $\Phi_{\text{o}}$, respectively.
Following 3DGS~\cite{kerbl3Dgaussians}, each 3D Gaussian is defined by its 3D position centroid and a set of Gaussian attributes~(\textit{e.g.}, scale, rotation, opacity, and appearance features) that jointly encode geometric and appearance properties.
To project 3D Gaussians onto the 2D image space, we adopt the differentiable rendering formulation introduced in Mip-Splatting~\cite{yu2024mip}.

\noindent\textbf{Human Gaussians.}
3D human Gaussians $\Phi_{\text{h}}$ are parameterized by Gaussian attributes $\phi_{\text{h}}$ along with SMPL-X human body pose $\theta$ and shape $\beta$ parameters.
The Gaussian attributes $\phi_{\text{h}}$ are defined in the canonical human pose, where each Gaussian is anchored to a surface point of the rest-posed SMPL-X~\cite{pavlakos2019expressive} mesh.
We then animate the Gaussians using linear blend skinning (LBS) driven by the human pose $\theta$ to obtain the final 3D human Gaussians.
In this animation process, we follow ExAvatar~\cite{moon2024expressive} to make the Gaussians of the hands and face follow the original SMPL-X skinning weights, while those of the other body parts adopt averaged skinning weights from their neighboring SMPL-X vertices.

\noindent\textbf{Object Gaussians.}
3D object Gaussians $\Phi_{\text{o}}$ are parameterized by Gaussian attributes~$\phi_{\text{o}}$, rotation~$R$, translation~$t$, and scale~$s$, where~$\phi_{\text{o}}$ is defined in a canonical space. 
The final 3D Gaussians are obtained by applying an affine transformation with rotation~$R$, translation~$t$, and scale~$s$ to the canonical representation.

\subsection{Reconstruction stage}
\label{sec:3.2_reconstruction_stage}
In this stage, we acquire the text prompts, the initial 3D human and object, and 2D background, which serve as necessary components for the HOI optimization stage.

\noindent\textbf{Text captioning.}
From the input image, we extract two text prompts, $P_{\text{holistic}}$ and $P_{\text{contact}}$.
$P_{\text{holistic}}$ captures the global context of the human–object interaction depicted in the image, while $P_{\text{contact}}$ specifies the human body parts (\textit{e.g.}, head and hands) involved in physical contact with the object.
These prompts are acquired using a vision-language model, GPT-4~\cite{achiam2023gpt}, which has powerful visual understanding capabilities learned from large-scale multimodal datasets.

\noindent\textbf{Human reconstruction.}
To obtain an initial 3D human reconstruction, we first remove the interacting object by using SmartEraser~\cite{jiang2025smarteraser}.
From the object-removed image, we segment the human region and mask the background to obtain a clean human image.
Based on the human segmented image, we derive the initial 3D Gaussian attributes $\phi_{\text{h}}$ with LHM~\cite{qiu2025lhm}.
Separately, we estimate the initial human pose $\theta$ and shape $\beta$ parameters using Multi-HMR~\cite{baradel2024multi}.

\noindent\textbf{Object reconstruction.}
To obtain an initial 3D object reconstruction, we also isolate the object from the input image with SmartEraser~\cite{jiang2025smarteraser} and SAM~\cite{kirillov2023segment}, acquiring a clean object image.
Based on the object image, we reconstruct its 3D shape and texture as a mesh with InstantMesh~\cite{xu2024instantmesh} and subsequently convert the mesh into 3D Gaussian attributes $\phi_{\text{o}}$ using 3DGS~\cite{kerbl3Dgaussians}.
The object pose parameters~($R$, $t$, and~$s$) are estimated by aligning the reconstructed 3D object surface with the depth map predicted by ZoeDepth~\cite{bhat2023zoedepth} from the original image.

\noindent\textbf{Background reconstruction.}
To build a 2D background, we utilize SmartEraser~\cite{jiang2025smarteraser} and remove the human and object from the input image.
The 2D background image, along with 3D human Gaussians~$\Phi_{\text{h}}$ and 3D object Gaussians~$\Phi_{\text{o}}$, are used for constructing realistic front-view (\textit{i.e.}, the input camera viewpoint) and novel-view renderings.

\subsection{HOI optimization stage}
\label{sec:3.3_hoi_optimization}
In this stage, we jointly refine the initial 3D human and object reconstructions using text prompts to ensure a holistic and semantically aligned reconstruction with the text.
Specifically, we optimize the 3D human Gaussians $\Phi_{\text{h}}$ and the 3D object Gaussians $\Phi_{\text{o}}$ over $N=200$ optimization steps, driven by the following overall loss function:
\begin{equation}
    \mathcal{L} = \mathcal{L}_\text{recon} + \mathcal{L}_\text{appr} + \mathcal{L}_\text{contact} + \mathcal{L}_\text{collision},
\end{equation}
where $\mathcal{L}_\text{collision}$ is a penalization term that discourages interpenetration between the 3D human and object, following Jiang~\etal~\cite{jiang2020coherent}.
The remaining terms are detailed below.

\noindent\textbf{Reconstruction loss.}
The reconstruction loss $\mathcal{L}_{\text{recon}}$ is defined as the discrepancy between the input image and the front-view rendering composed of 3D human Gaussians, 3D object Gaussians, and 2D background.
It consists of two mean squared error (MSE) terms: (1) loss between the rendered RGB image and the input image, and (2) loss between the rendered silhouettes and the corresponding segmentation masks of the human and object in the input image.

\noindent\textbf{Appearance loss.}
The appearance loss $\mathcal{L}_{\text{appr}}$ is defined as the semantic distance between the holistic text prompt $P_{\text{holistic}}$ and the novel-view rendering composed of 3D human Gaussians, 3D object Gaussians, and the 2D background.
Specifically, we sample a random viewpoint uniformly over a sphere, render the 3D human and object Gaussians from that viewpoint, and composite the result onto the 2D background image.
Based on the rendered image, we apply the score distillation sampling strategy~\cite{poole2022dreamfusion} by leveraging the rich visual prior knowledge of a pre-trained diffusion network (\textit{e.g.}, StableDiffusion~\cite{rombach2022high}) to align the rendered appearance with the text prompt.
The appearance loss is computed as
\begin{equation}
\label{eq:appr_loss}
\begin{split}
\nabla_{\Phi} \mathcal{L}_{\text{appr}} = \mathbb{E}[w_t (\hat{\epsilon}_{t}(\mathbf{x}_{t}; P_{\text{holistic}}) - \epsilon_{t}) \frac{\partial \mathbf{x}_{t}}{\partial \Phi}],
\end{split}
\end{equation} 
where $t$ denotes the noise level, $\mathbf{x}_{t}$ is the rendered image perturbed by Gaussian noise $\epsilon_{t}$, and $w_{t}$ is a weighting factor determined by the noise level $t$.
The loss function minimizes the discrepancy between the predicted noise $\hat{\epsilon}_{t}(\cdot)$ and the true noise $\epsilon_{t}$.
This encourages the optimization of the 3D Gaussians, through the rendered images, to align with the distribution of plausible human–object interaction appearances learned by the pre-trained diffusion network.

The appearance loss effectively utilizes the visual prior knowledge from the pre-trained diffusion network to guide the reconstruction toward plausible human-object interactions, addressing the absence of interaction reasoning in the initial 3D human and object reconstruction.
It captures holistic contextual cues beyond physical contact, enabling reasoning about non-contact interactions and object orientation related to human intent.
By applying such holistic guidance, it complements the contact information and significantly improves the accuracy and plausibility of 3D human and object reconstruction.

\begin{figure}[t!]
  \vspace*{+0.1em}
  \centering
  \includegraphics[width=1.0\linewidth]{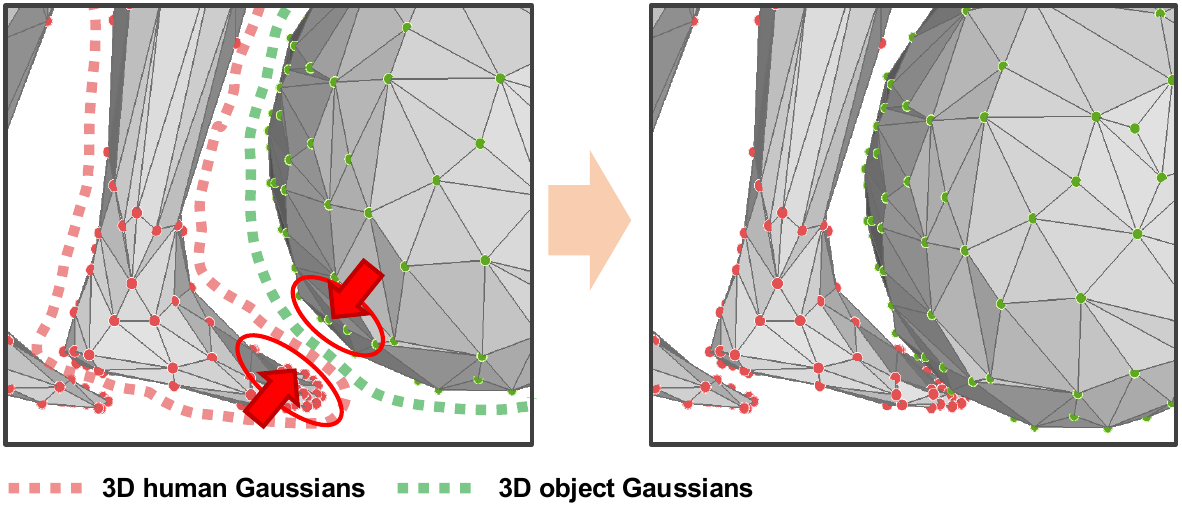}
  \vspace*{-1.7em}
  \caption{\textbf{Gaussians-to-mesh conversion process.}
  }
  \vspace*{+0.0em}
  \label{fig:3_smplx_shift}
\end{figure}

\noindent\textbf{Contact loss.}
The contact loss $\mathcal{L}_{\text{contact}}$ is defined as the proximity between the object surface and human body parts predicted to be in contact.
We identify the set of Gaussian center points~$V_{h,c}$ corresponding to the contacting body parts specified in the SMPL-X body model, based on the contact text prompt $P_{\text{contact}}$ (\textit{e.g.}, right foot).
Then, we compute the loss that minimizes the distance between the contacting 3D human points $V_{h,c}$ and their nearest 3D object points $V_o$, within a threshold $\tau = 10~\text{cm}$, calculated as:
\begin{equation}
    \mathcal{L}_{\text{contact}} = \frac{1}{\vert V_{h,c} \vert} \sum_{v_h \in V_{h,c}} d(v_h, V_o) \cdot \mathbbm{1}[d(v_h, V_o) < \tau].
\end{equation}
This loss enforces local physical plausibility between the estimated contacting regions, enhancing reconstruction accuracy in conjunction with the appearance loss $\mathcal{L}_{\text{appr}}$.

\subsection{Gaussians-to-mesh conversion}
\label{sec:3.4_gaussian_to_mesh}
\cref{fig:3_smplx_shift} illustrates the process of converting the 3D Gaussians of the final human and object reconstruction into 3D human and object meshes.
Naturally, the 3D Gaussians deviate from their underlying base meshes, which are the SMPL-X body model~\cite{pavlakos2019expressive} for the human and the 3D object mesh from InstantMesh~\cite{xu2024instantmesh}.
For contact regions, direct conversion from 3D Gaussians to a mesh can result in inconsistencies between the contacts defined by the 3D Gaussians and those defined by the corresponding mesh surfaces.
To ensure consistent contact, we apply a local shift that moves the object mesh toward the human surface to resolve such inconsistencies.
To this end, we identify contact regions where the distance between the 3D human and object Gaussians is less than $5~\text{cm}$.
Then, we select the mesh vertices on both the human and object surfaces that correspond to these contact regions and minimize the distance between these vertices to zero.
We use the 3D human and object meshes after this conversion in all experiments described in \cref{sec:experiments}, for comparison with existing mesh-based reconstruction methods.

\section{Experiments}
\label{sec:experiments}
\subsection{Datasets}
Open3DHOI~\cite{wen2025reconstructing} and BEHAVE~\cite{bhatnagar2022behave} datasets are used for our experiments.
Open3DHOI is an open-vocabulary, in-the-wild 3D HOI dataset that we use only for evaluation. 
It contains over 2.5K images and 133 object categories.
BEHAVE is an indoor 3D HOI dataset that captures the interactions of 8 human subjects and 20 objects in a controlled setting.
We use its official test set, which consists of 4.5K images, for the evaluation.

\subsection{Evaluation metrics}
\label{sec:evaluation_metrics}
For all evaluation metrics, we align the root position between the 3D human reconstruction and ground-truth (GT).

\noindent\textbf{Chamfer distance ($\text{CD}_{\text{human}}$, $\text{CD}_{\text{object}}$).}
We evaluate the 3D human and object reconstruction using the Chamfer distance between the predicted 3D surface and the corresponding GT.
The Chamfer distance is separately computed for the human ($\text{CD}_{\text{human}}$) and object ($\text{CD}_{\text{object}}$) in centimeters.

\noindent\textbf{Contact score.}
We evaluate contact fidelity between reconstructed human and object using the F1 score of contact regions derived from their 3D surfaces.
We obtain a contact map by extracting human vertices of the SMPL-X mesh surface~\cite{pavlakos2019expressive} within $5~\text{cm}$ of the object mesh.
Then, the contact score ($\text{Contact}$) is computed as the harmonic mean of precision and recall, defined as $2 \times \frac{PR}{(P + R)}$, where $P$ and $R$ denote precision and recall.
Note that contact evaluation is only performed on the SMPL-X mesh surface, since GT contacts are defined on the SMPL-X topology and are incompatible with the topology-free Gaussian representation.

\noindent\textbf{Collision.}
To evaluate physical plausibility, we evaluate collision by measuring the interpenetration between the reconstructed 3D human and object.
We compute the percentage of human vertices located within the object mesh.

\subsection{Ablation study}
We carry out all ablation studies on Open3DHOI~\cite{wen2025reconstructing}.

\begin{figure}[t!]
  \centering
  \includegraphics[width=1.0\linewidth]{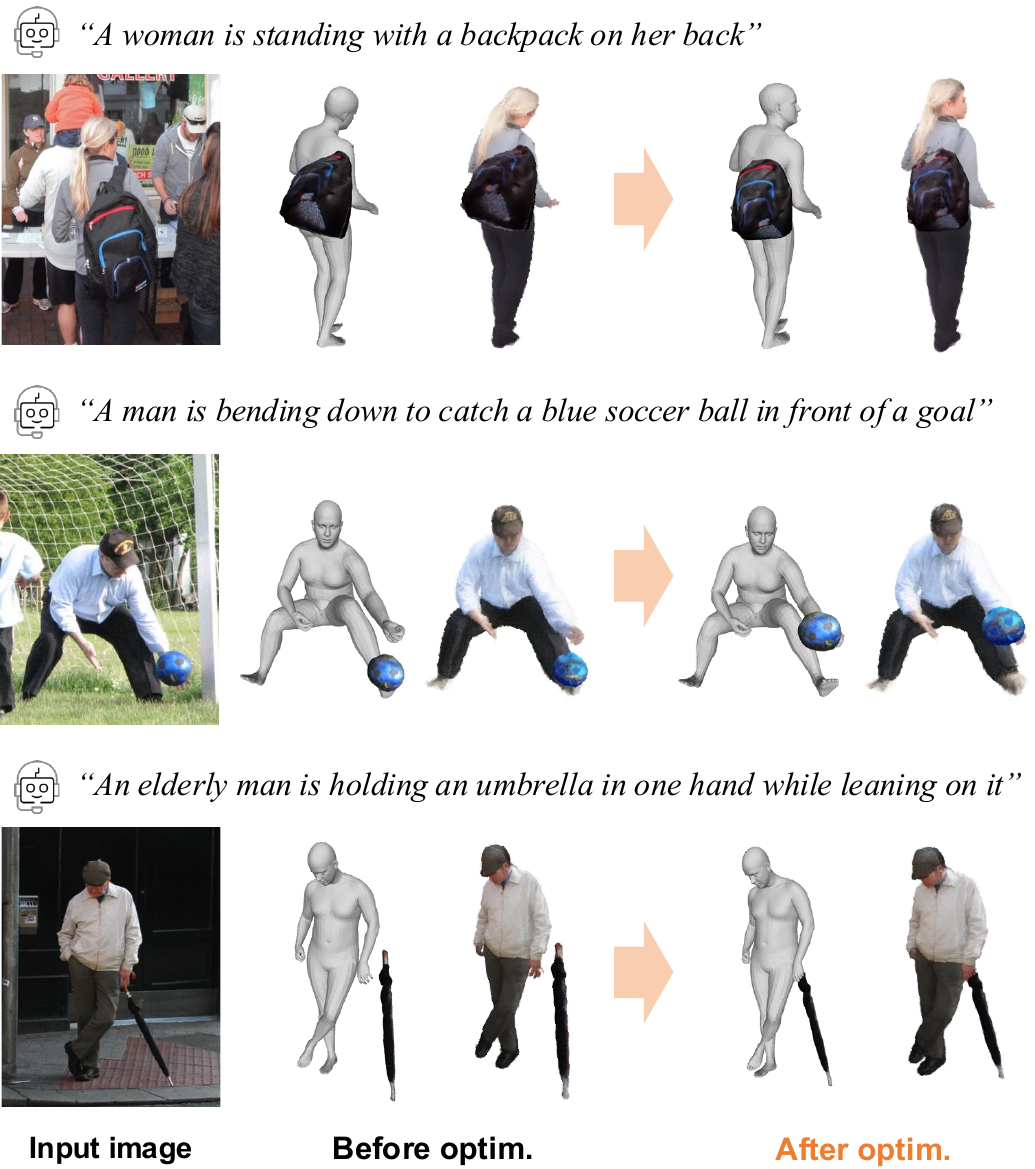}
  \vspace*{-1.4em}
  \caption{\textbf{Optimization results of \ourmethod.}
  Our text-guided optimization accurately refines the 3D human and object by utilizing their corresponding text descriptions.
  }
  \vspace*{+0.0em}
  \label{fig:4_optimization_results}
\end{figure}

\begin{figure}[t!]
  \centering
  \includegraphics[width=1.0\linewidth]{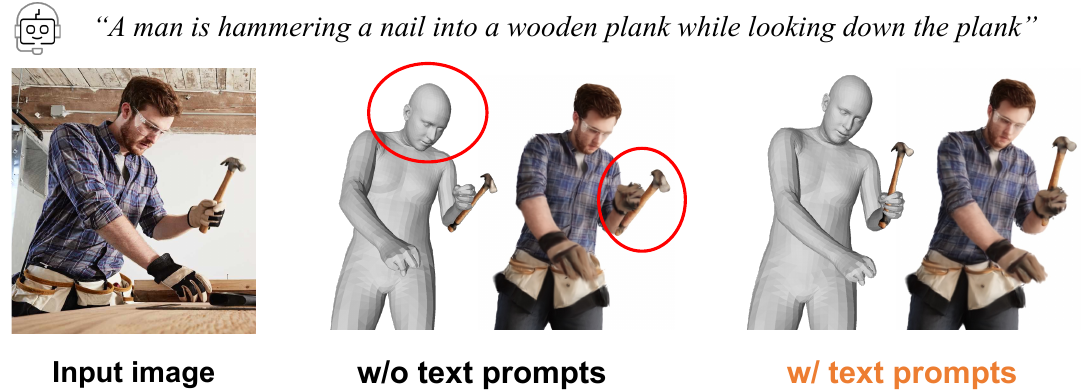}
  \vspace*{-1.3em}
  \caption{\textbf{Effectiveness of text descriptions in optimization.}
  }
  \vspace*{-0.0em}
  \label{fig:5_ablation_text}
\end{figure}

\noindent\textbf{Effectiveness of text-guided optimization.}
\cref{fig:4_optimization_results} and \cref{tab:abl_optimization} demonstrate that our optimization stage effectively refines the 3D human and object by capturing semantic interactions described in text descriptions.
When we remove the text prompt condition from the appearance loss $\mathcal{L}_{\text{appr}}$, the reconstruction fails to capture the global context of the human-object interaction.
As shown in \cref{fig:5_ablation_text}, optimization without text conditioning fails to orient the human’s gaze toward the right hand, whereas incorporating text information corrects this implicit context, resulting in accurate 3D human and object reconstruction.
This is mainly because the text descriptions provide the holistic context about the interaction beyond the physical contact information.
Moreover, the texts provide the semantic prior knowledge for inferring non-contact interactions, which cannot be derived from the contact information.
Thus, leveraging the comprehensive textual information enables our framework to produce accurate and plausible reconstructions that capture the global interaction context.

\begin{figure}[t!]
  \centering
  \includegraphics[width=1.0\linewidth]{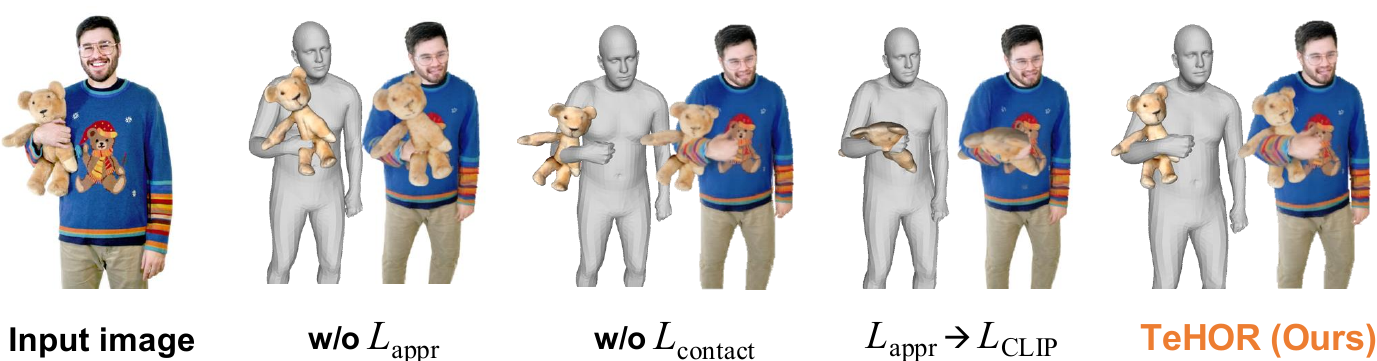}
  \vspace*{-1.4 em}
  \caption{\textbf{Effectiveness of each loss function in our framework.}
  }
  \vspace*{+0.0em}
  \label{fig:6_ablation_loss}
\end{figure}

\noindent\textbf{Ablation on loss configurations.} 
\cref{fig:6_ablation_loss} and \cref{tab:abl_loss} demonstrate the effectiveness of our loss design, comprising appearance loss $\mathcal{L}_{\text{appr}}$ and contact loss $\mathcal{L}_{\text{contact}}$.
The first row of the table shows that removing the appearance loss fails to capture holistic interaction cues, leading to implausible reconstructions.
As shown in \cref{fig:6_ablation_loss}, the teddy bear held by the person is displaced from the chest and even penetrates the arm.
With the inclusion of appearance loss, contextual cues from the input image are effectively captured and reflected in the 3D reconstruction, resulting in a consistent human–object interaction.
To further examine the effectiveness of our appearance loss, we replace the appearance loss~$\mathcal{L}_{\text{appr}}$ with CLIP loss~$\mathcal{L}_{\text{CLIP}}$, which also provides appearance-level supervision using the textual information. 
Specifically, the CLIP loss function minimizes the cosine distance between the feature embeddings of the rendered image and the corresponding text prompt within the pre-trained CLIP embedding space~\cite{radford2021learning}.
As shown in the third and fourth rows of \cref{tab:abl_loss}, our appearance loss achieves significantly better object Chamfer distance~($\text{CD}_{\text{object}}$) and contact F1-score.
CLIP encodes text features into a single 1D embedding vector, which limits its ability to model dense spatial human-object relationships.
In contrast, our appearance loss $\mathcal{L}_{\text{appr}}$ applies dense, pixel-level gradients in rendered 2D appearances, distilling rich priors from the pre-trained diffusion network to enable accurate, fine-grained 3D reconstruction.
Accordingly, our appearance loss effectively optimizes the 3D human and object to follow the appearance cues under the diffusion prior, ensuring both plausible and accurate 3D reconstruction.

\begin{table}[t]
\def\arraystretch{1.8}
\renewcommand{\tabcolsep}{0.8mm}
\footnotesize
\begin{center}
\scalebox{0.8}{
    \begin{tabular}{>{\raggedright\arraybackslash}m{3.8cm}|>{\centering\arraybackslash}m{1.35cm}>{\centering\arraybackslash}m{1.35cm}>{\centering\arraybackslash}m{1.35cm}>{\centering\arraybackslash}m{1.35cm}}
    \specialrule{.1em}{.05em}{0.0em}
         & CD$_{\text{human}}{\downarrow}$ & CD$_{\text{object}}{\downarrow}$ & Contact${\uparrow}$ & Collision${\downarrow}$  \\ \hline
        Before optim. & 5.252 & 31.268 & 0.305 & \textbf{0.040} \\
        After optim. w/o text prompts & 5.028 & 20.348  & 0.374 & 0.052 \\
        \textbf{After optim. (Ours)} & \textbf{4.941} & \textbf{16.701} & \textbf{0.412} & 0.047 \\
        \specialrule{.1em}{-0.05em}{-0.05em}
    \end{tabular}
    \vspace*{-1mm}
}
\end{center}
    \vspace*{-1.4em}
    \caption{
    \textbf{Effectiveness of text-guided optimization.}
    }
    \vspace*{+0.0em}
    \label{tab:abl_optimization}
\end{table}

\noindent\textbf{Ablation on appearance rendering.}
\cref{tab:abl_components} investigates the impact of key components in our framework that enable realistic appearance rendering. 
We first evaluate the advantage of the 3D Gaussian representation over mesh, which is a widely used representation in 3D human and object reconstruction~\cite{zhang2020perceiving,cseke2025pico,dwivedi2025interactvlm}.
As shown in the first row, 3D Gaussians significantly outperform mesh representation for both the human and object, due to two main factors.
First, 3D Gaussians excel at modeling high-fidelity visual appearances, providing richer signals that enable the appearance loss $\mathcal{L}_{\text{appr}}$ to more accurately align the reconstruction with the textual semantics.
Second, their flexible and topology-free structure allows for more effective optimization of the 3D spatial relationships between the human and object.
We also conduct an ablation study on the use of a 2D background for realistic rendering.
The second row of \cref{tab:abl_components} shows a significant degradation in performance when the 2D background is removed.
This demonstrates that the 2D background plays a crucial role in providing complete scene context, which allows the appearance loss to fully exploit the image prior knowledge from the diffusion network, leading to more precise optimization.

\subsection{Comparison with state-of-the-art methods}
We compare our framework with state-of-the-art 3D human and object reconstruction methods.
Since PICO~\cite{cseke2025pico} requires human–object contact as input, we run the method using contact estimation results from LEMON~\cite{yang2024lemon}.
For a fair comparison, we use the same 3D human and object initializations for all methods.
Specifically, the initial 3D human pose is obtained from Multi-HMR~\cite{baradel2024multi}, and the initial 3D object pose is estimated by aligning the reconstructed 3D object with the depth map predicted by ZoeDepth~\cite{bhat2023zoedepth}.
All object shapes are reconstructed using InstantMesh~\cite{xu2024instantmesh}.
As \ourmethod~is a Gaussian-based framework, we convert the initial object mesh into 3D Gaussians via 3DGS~\cite{kerbl3Dgaussians}.
The $\dagger$ symbol denotes the evaluation that directly compares the reconstructed 3D Gaussians against the GT mesh surfaces, without Gaussian-to-mesh conversion.

\begin{table}[t]
\def\arraystretch{1.78}
\renewcommand{\tabcolsep}{0.8mm}
\footnotesize
\begin{center}
\scalebox{0.83}{
    \begin{tabular}{>{\centering\arraybackslash}m{1.55cm}>{\centering\arraybackslash}m{1.55cm}|>{\centering\arraybackslash}m{1.35cm}>{\centering\arraybackslash}m{1.35cm}>{\centering\arraybackslash}m{1.35cm}>{\centering\arraybackslash}m{1.35cm}}
    \specialrule{.1em}{.05em}{0.0em}
        $L_{\text{appr}}$ & $L_{\text{contact}}$ & CD$_{\text{human}}{\downarrow}$ & CD$_{\text{object}}{\downarrow}$ & Contact${\uparrow}$ & Collision${\downarrow}$  \\
        \hline
        \xmark & \cmark & 5.191 & 22.094 & 0.330 & 0.049 \\
        \cmark & \xmark & 5.311 & 19.849 & 0.374 & 0.054 \\
        {\fontsize{8}{16}\selectfont $L_{\text{CLIP}}$} & \cmark & 5.018 & 18.504 & 0.366 & \textbf{0.047} \\
        \textbf{\cmark\makebox[0pt][l]{\quad(Ours)}} & \textbf{\cmark}& \textbf{4.941} & \textbf{16.701} & \textbf{0.412} & \textbf{0.047} \\
        \specialrule{.1em}{-0.05em}{-0.05em}
    \end{tabular}
    \vspace*{-1mm}
}
\end{center}
    \vspace*{-1.4em}
    \caption{
    \textbf{Ablation studies for loss configurations.}
    }
    \vspace*{+0.0em}
    \label{tab:abl_loss}
\end{table}
\begin{table}[t]
\def\arraystretch{1.78}
\renewcommand{\tabcolsep}{0.8mm}
\footnotesize
\begin{center}
\scalebox{0.83}{
    \begin{tabular}{>{\raggedright\arraybackslash}m{3.1cm}|>{\centering\arraybackslash}m{1.35cm}>{\centering\arraybackslash}m{1.35cm}>{\centering\arraybackslash}m{1.35cm}>{\centering\arraybackslash}m{1.35cm}}
    \specialrule{.1em}{.05em}{0.0em}
         & CD$_{\text{human}}{\downarrow}$ & CD$_{\text{object}}{\downarrow}$  & Contact${\uparrow}$ & Collision${\downarrow}$  \\ \hline
        3D Gaussians {\fontsize{8}{16}\selectfont $\rightarrow$} Mesh & 5.153 & 25.162 & 0.308 & 0.054 \\
        w/o 2D background & 5.002 & 18.196 & 0.389 & 0.049 \\
        \textbf{\ourmethod~(Ours)} & \textbf{4.941} & \textbf{16.701} & \textbf{0.412} & \textbf{0.047} \\
        \specialrule{.1em}{-0.05em}{-0.05em}
    \end{tabular}
    \vspace*{-1mm}
}
\end{center}
    \vspace*{-1.4em}
    \caption{
    \textbf{Ablation studies for appearance rendering.}
    }
    \vspace*{+0.0em}
    \label{tab:abl_components}
\end{table}

\begin{figure*}[t!]
  \centering
  \includegraphics[width=0.97\linewidth]{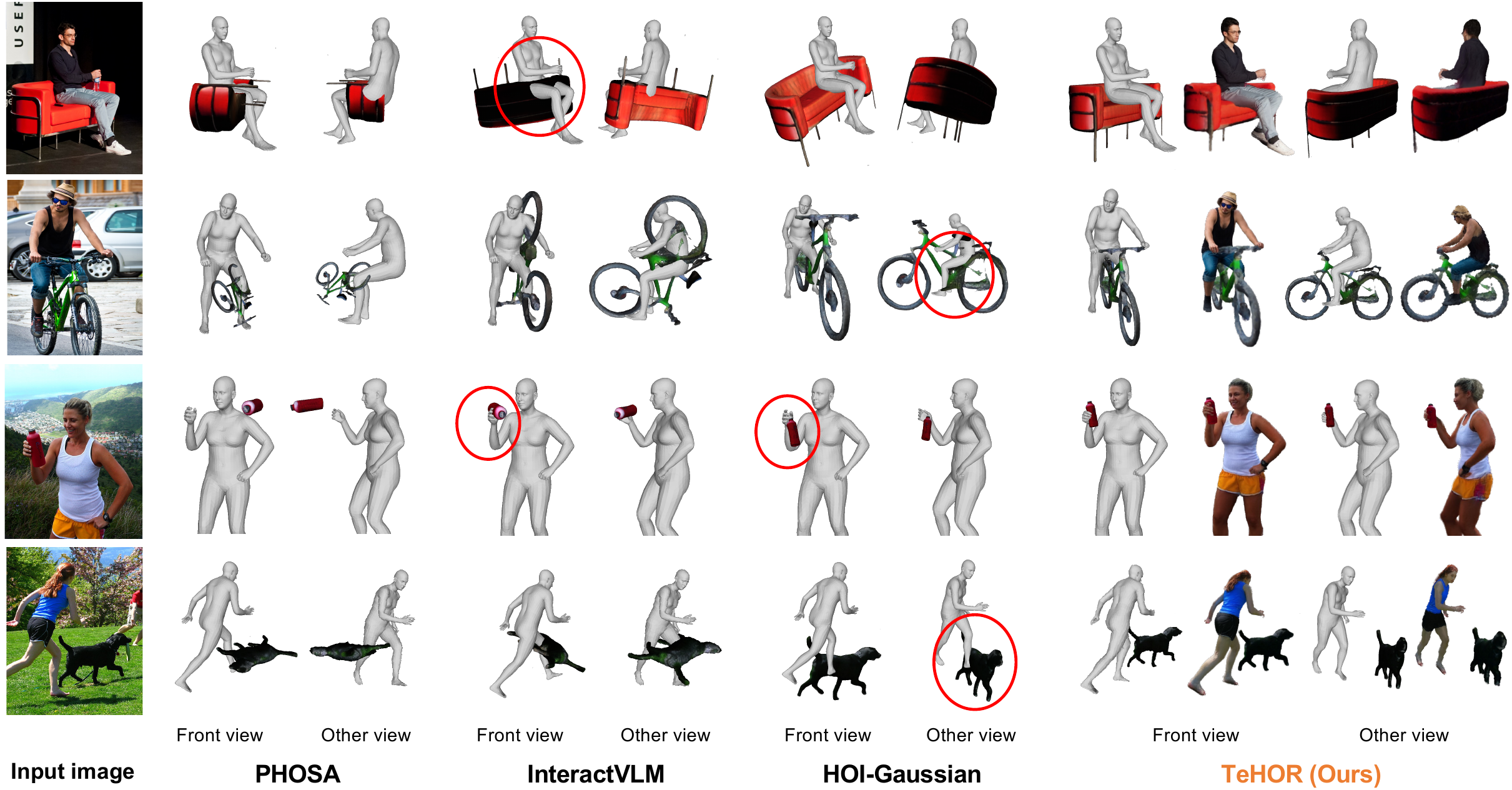}
  \vspace*{-0.4em}
  \caption{\textbf{Qualitative comparison with state-of-the-art methods.}
  We highlight their representative failure cases with red circles. 
  }
   \vspace*{+0.0em}
  \label{fig:7_comparison_sota}
\end{figure*}
\begin{table*}[t]
\def\arraystretch{1.5}
\renewcommand{\tabcolsep}{0.8mm}
\footnotesize
\begin{center}
\scalebox{0.8}{
    \begin{tabular}{>{\raggedright\arraybackslash}m{4.4cm}|>{\centering\arraybackslash}m{1.82cm}>{\centering\arraybackslash}m{1.82cm}>{\centering\arraybackslash}m{1.82cm}>{\centering\arraybackslash}m{1.95cm}|>{\centering\arraybackslash}m{1.82cm}>{\centering\arraybackslash}m{1.82cm}>{\centering\arraybackslash}m{1.82cm}>{\centering\arraybackslash}m{1.95cm}}
    \specialrule{.1em}{.05em}{0.0em}
        &  \multicolumn{4}{c|}{Open3DHOI} & \multicolumn{4}{c}{BEHAVE} \\
        Methods  & CD$_{\text{human}}{\downarrow}$ & CD$_{\text{object}}{\downarrow}$ & Contact${\uparrow}$ & Collision${\downarrow}$ & CD$_{\text{human}}{\downarrow}$ & CD$_{\text{object}}{\downarrow}$ & Contact${\uparrow}$ & Collision${\downarrow}$  \\
        \hline
        PHOSA~\cite{zhang2020perceiving} & 5.342 & 49.180 & 0.243 & \textbf{0.044} & 5.758 & 46.003 & 0.257 & \textbf{0.010} \\ 
        LEMON~\cite{yang2024lemon} + PICO~\cite{cseke2025pico} & 5.948 & 25.889 & 0.335 & 0.078 & 6.159 & 22.585 & 0.082 & 0.045 \\
        InteractVLM~\cite{dwivedi2025interactvlm} & 5.252 & 24.238 & \underline{0.392} & 0.054 & 5.770 & \underline{19.197} & \underline{0.379} & 0.021 \\
        HOI-Gaussian~\cite{wen2025reconstructing} & \underline{5.111} & \underline{19.363} & 0.348 & 0.070 & \underline{5.748} & 21.774 & 0.371 & 0.019 \\
        \textbf{\ourmethod~(Ours)} & \textbf{4.941} & \textbf{16.701} & \textbf{0.412} & \underline{0.047} & \textbf{5.615} & \textbf{17.339} & \textbf{0.412} & \underline{0.016} \\ \cdashline{1-9}
        \textbf{\ourmethod$^{\dagger}$~(Ours)} & 4.403 & 16.697 & -- & 0.045 & 5.241 & 17.341 & -- & 0.012 \\
        \specialrule{.1em}{-0.05em}{-0.05em}
    \end{tabular}
}
\end{center}
    \vspace*{-1.4em}
    \caption{
        \textbf{Quantitative comparison with state-of-the-art methods on Open3DHOI~\cite{wen2025reconstructing} and BEHAVE~\cite{bhatnagar2022behave}.}
        Bold and underlined values indicate the best and second-best scores, respectively.
        $\dagger$ denotes the evaluation results directly computed based on the centers of reconstructed 3D Gaussians instead of 3D mesh vertices.
    }
    \vspace*{-0.5em}
    \label{tab:sota_comparison}
\end{table*}
\begin{table}[t]
\vspace*{+0.5em}
\def\arraystretch{1.6}
\renewcommand{\tabcolsep}{0.8mm}
\footnotesize
\begin{center}
\scalebox{0.8}{
    \begin{tabular}{>{\raggedright\arraybackslash}m{3.6cm}|>{\centering\arraybackslash}m{1.60cm}>{\centering\arraybackslash}m{1.60cm}>{\centering\arraybackslash}m{1.60cm}}
    \specialrule{.1em}{.05em}{0.0em}
         Methods & CD$_{\text{human}}{\downarrow}$ & CD$_{\text{object}}{\downarrow}$ & Collision${\downarrow}$  \\ \hline
        PHOSA~\cite{zhang2020perceiving} & 5.401 & 65.537 & 0.028  \\
        LEMON~\cite{yang2024lemon} + PICO~\cite{cseke2025pico} & 5.635 & 33.073 & 0.029 \\
        InteractVLM~\cite{dwivedi2025interactvlm} & 5.390 & 46.819 & 0.011 \\
        HOI-Gaussian~\cite{wen2025reconstructing} & 5.244 & 25.374 & 0.037 \\
        \textbf{\ourmethod~(Ours)} & \textbf{4.958} & \textbf{17.546} & \textbf{0.005} \\
        \specialrule{.1em}{-0.05em}{-0.05em}
    \end{tabular}
    \vspace*{-1mm}
}
\end{center}
    \vspace*{-1.4em}
    \caption{
    \textbf{Quantitative comparison with state-of-the-art methods for non-contact scenarios on Open3DHOI~\cite{wen2025reconstructing}.}
    }
    \vspace*{-0.5em}
    \label{tab:non_contact}
\end{table}

\cref{fig:7_comparison_sota} and \cref{tab:sota_comparison} show that our \ourmethod~largely outperforms all state-of-the-art methods both qualitatively and quantitatively.
As previous methods rely heavily on local geometric cues, such as human–object contact information, they often fail to handle interactions that require global context, including object orientation.
Although HOI-Gaussian~\cite{wen2025reconstructing} uses depth map from the input image as additional information for the reconstruction, it is unable to capture the semantic context of the interaction and is vulnerable to severe human-object occlusion.
Unlike the previous methods, our \ourmethod~leverages rich text descriptions of human–object interactions as a key prior, providing holistic and semantic context that contact information alone cannot capture.
Additionally, \cref{tab:non_contact}~demonstrates our superiority in non-contact scenarios, where we evaluate the methods on a subset of Open3DHOI that excludes samples with physical contact between the ground-truth 3D human and object.
Such cases are particularly challenging because contact-based cues vanish entirely, forcing the reconstruction system to reason about the interaction from global contextual signals such as object orientation, gaze direction, and body posture.
While previous methods fail to reason about non-contact interaction, our framework benefits from text descriptions, achieving superior performance in the non-contact scenarios.
Overall, our framework, which effectively optimizes the 3D human and object using these comprehensive text descriptions, provides more accurate and robust reconstructions than previous methods.
\section{Conclusion}
We propose~\textbf{\ourmethod}, a text-guided framework for joint reconstruction of the 3D human and object from a single image.
Our framework leverages text descriptions to enforce semantic alignment, enabling reasoning over a wide spectrum of interactions, including non-contact cases.
It incorporates appearance cues from the 3D human and object to capture a holistic context, ensuring visual plausibility of the reconstruction.
Extensive experiments demonstrate that our framework produces accurate and semantically plausible reconstructions, achieving state-of-the-art performance.

\vspace*{+0.7em}
\noindent\textbf{Acknowledgements.}
This work was supported in part by the IITP grants [No. RS-2021-II211343, Artificial Intelligence Graduate School Program (Seoul National University), No. RS-2025-02303870, No. 2022-0-00156] funded by the Korea government (MSIT).

\clearpage
\setcounter{page}{1}
\maketitlesupplementary
\vspace*{+1.5em}

\setcounter{section}{0}
\setcounter{table}{0}
\setcounter{figure}{0}
\renewcommand{\thesection}{S\arabic{section}}   
\renewcommand{\thetable}{S\arabic{table}}   
\renewcommand{\thefigure}{S\arabic{figure}}

In this supplementary material, we present additional technical details and more experimental results that could not be included in the main manuscript due to the lack of pages.
The contents are summarized below:
\begin{itemize}
\vspace{2.5mm}
\item \ref{sec:suppl_text_align}. Evaluation on semantic alignment
\item \ref{sec:suppl_hdm}. Comparison with HDM
\item \ref{sec:suppl_contact}. Impact of contact estimation accuracy
\item \ref{sec:suppl_gs_optim}. Impact of Gaussian attributes optimization
\item \ref{sec:suppl_g2m}. Impact of Gaussians-to-mesh conversion
\item \ref{sec:suppl_captioning}. Details of text captioning
\item \ref{sec:suppl_implementation}. Implementation details
\item \ref{sec:suppl_limitation}. Limitations and future work
\item \ref{sec:suppl_more_results}. More qualitative results
\end{itemize}


\section{Evaluation on semantic alignment}
\label{sec:suppl_text_align}
\begin{figure}[t!]
  \vspace*{+2.7em}
  \centering
  \includegraphics[width=1.0\linewidth]{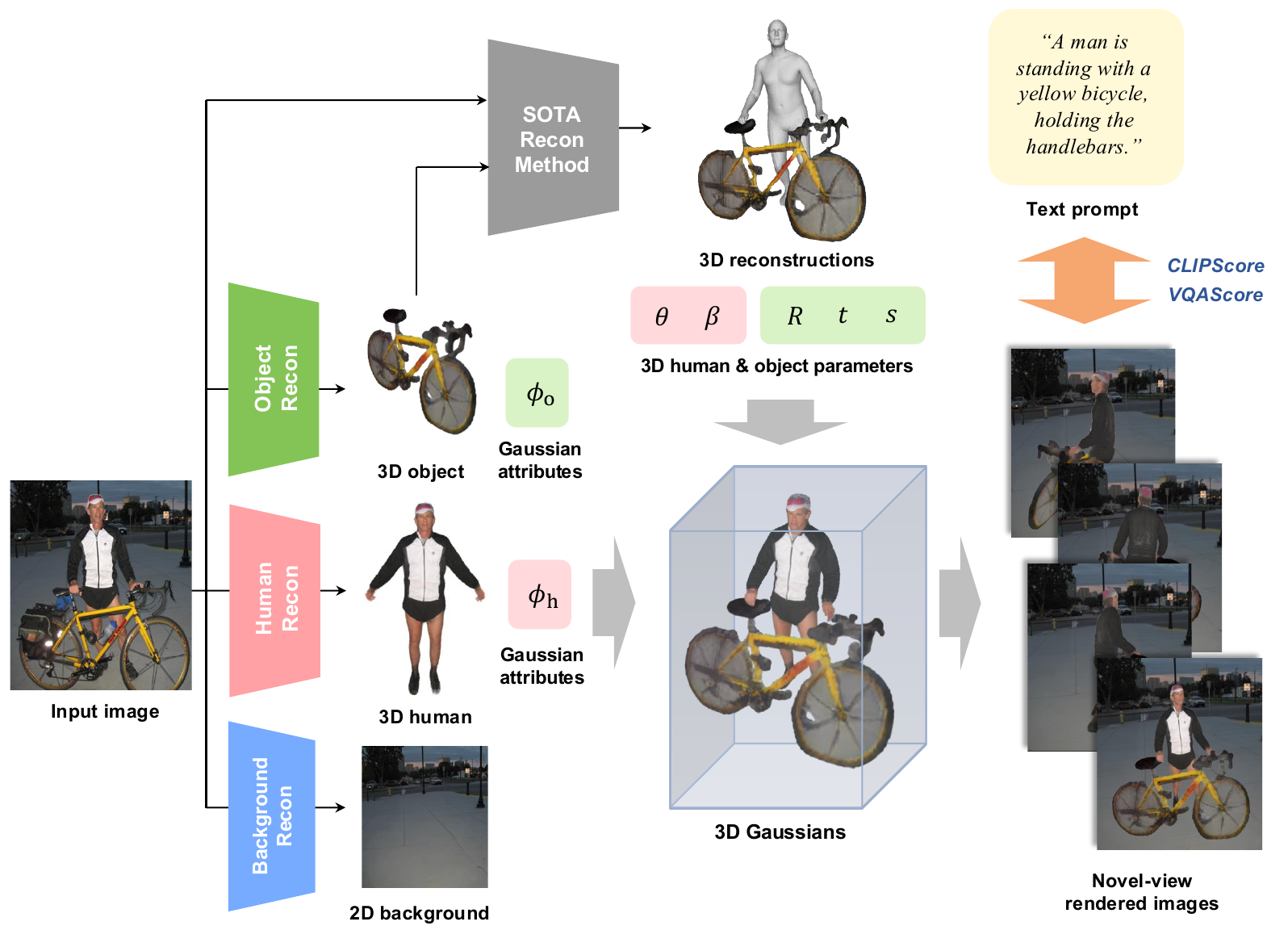}
  \vspace*{-1.4em}
  \caption{\textbf{Process of evaluating text alignment for state-of-the-art reconstruction methods.}
  }
  \vspace*{+0.0em}
  \label{fig:suppl_text_align}
\end{figure}

In this section, we introduce additional evaluation to compare \ourmethod~with state-of-the-art reconstruction methods on semantic alignment between 3D reconstructions and text descriptions.
Since direct comparison between 3D reconstructions and text is not feasible, we instead evaluate appearance-text alignment metrics on 2D renderings of the reconstructions, following the process shown in \cref{fig:suppl_text_align}.
To ensure a fair comparison, we unify the underlying 3D representation across all methods, since existing methods predominantly use mesh-based representations, whereas our framework is Gaussian-based.
Specifically, we use the same initial 3D human and object Gaussians, including both shape and texture attributes, for all methods.
We then extract each method's human ($\theta$ and $\beta$) and object ($R$, $t$, and $s$) pose parameters and apply them to transform the 3D Gaussians.
This setup ensures that the only variable in the experiments is the set of 3D pose parameters provided by each method.
For evaluation, we render the transformed Gaussians on the 2D background from pre-defined viewpoints 0$^{\circ}$, 90$^{\circ}$, 180$^{\circ}$, and 270$^{\circ}$.
Each rendered image is then paired with its corresponding text description to compute two image-text alignment metrics: 1) CLIPScore~\cite{hessel2021clipscore} and 2) VQAScore~\cite{lin2024evaluating}.
CLIPScore computes the cosine similarity between the embeddings of the rendered image and the text description.
VQAScore utilizes a powerful visual-question-answering (VQA) model to compute the alignment score by converting the text description into a simple query and measuring the generative likelihood of a desired response.
Here, we use InstructBLIP-FlanT5-XL~\cite{dai2023instructblip} as the underlying VQA model to compute VQAScore.
\cref{tab:suppl_text_align} shows that our framework outperforms other state-of-the-art methods in text alignment by effectively capturing the holistic and semantic context of human-object interaction.

\begin{table}[t]
\def\arraystretch{1.52}
\renewcommand{\tabcolsep}{0.8mm}
\footnotesize
\begin{center}
\scalebox{0.77}{
    \begin{tabular}{>{\raggedright\arraybackslash}m{3.85cm}|>{\centering\arraybackslash}m{2.20cm}>{\centering\arraybackslash}m{2.20cm}}
    \specialrule{.1em}{.05em}{0.0em}
         &  \multicolumn{2}{c}{Open3DHOI} \\
         Methods & CLIPScore${\uparrow}$ & VQAScore${\uparrow}$ \\ \hline
        PHOSA~\cite{zhang2020perceiving} & 0.689 & 0.631 \\
        LEMON~\cite{yang2024lemon} + PICO~\cite{cseke2025pico} & 0.696 & 0.642 \\
        InteractVLM~\cite{dwivedi2025interactvlm} & 0.694 & 0.647 \\
        HOI-Gaussian~\cite{wen2025reconstructing} & 0.698 & 0.648 \\
        \textbf{\ourmethod~(Ours)} & \textbf{0.706} & \textbf{0.652} \\
        \specialrule{.1em}{-0.05em}{-0.05em}
    \end{tabular}
    \vspace*{-1mm}
}
\end{center}
    \vspace*{-1.2em}
    \caption{
    \textbf{Quantitative evaluation of appearance-text alignment.}
    }
    \vspace*{-0.2em}
    \label{tab:suppl_text_align}
\end{table}

\begin{table*}[t!]
\def\arraystretch{1.55}
\renewcommand{\tabcolsep}{0.8mm}
\footnotesize
\begin{center}
\scalebox{0.72}{
    \begin{tabular}{>{\raggedright\arraybackslash}m{3.9cm}|>{\centering\arraybackslash}m{2.2cm}>{\centering\arraybackslash}m{2.2cm}>{\centering\arraybackslash}m{2.2cm}|>{\centering\arraybackslash}m{2.2cm}>{\centering\arraybackslash}m{2.2cm}>{\centering\arraybackslash}m{2.2cm}}
    \specialrule{.1em}{.05em}{0.0em}
        &  \multicolumn{3}{c|}{Seen object categories} & \multicolumn{3}{c}{Whole object categories} \\ [-0.5ex]
        & CD$_{\text{human}}{\downarrow}$ & CD$_{\text{object}}{\downarrow}$ & Collision${\downarrow}$ & CD$_{\text{human}}{\downarrow}$ & CD$_{\text{object}}{\downarrow}$ & Collision${\downarrow}$ \\ \hline
        HDM~\cite{xie2023templatefree} & 4.977 & 16.349 & 0.040 & 6.084 & 30.422 & \textbf{0.038} \\
        \textbf{TeHOR (Ours)} & \textbf{2.511} & \textbf{14.905} & \textbf{0.035} & \textbf{2.582} & \textbf{17.938} & 0.047 \\
        \specialrule{.1em}{-0.05em}{-0.05em}
      \end{tabular}
    \vspace*{-1mm}
}
\end{center}
    \vspace*{-1.5em}
    \caption{
    \textbf{Quantitative comparison with HDM~\cite{xie2023templatefree} on Open3DHOI~\cite{wen2025reconstructing}.}
    }
    \vspace*{-0.3em}
    \label{tab:suppl_hdm}
\end{table*}
\begin{table*}[t!]
\def\arraystretch{1.55}
\renewcommand{\tabcolsep}{0.8mm}
\footnotesize
\begin{center}
\scalebox{0.72}{
    \begin{tabular}{>{\raggedright\arraybackslash}m{4.0cm}|>{\centering\arraybackslash}m{1.90cm}>{\centering\arraybackslash}m{1.90cm}>{\centering\arraybackslash}m{1.90cm}|>{\centering\arraybackslash}m{1.8cm}>{\centering\arraybackslash}m{1.80cm}>{\centering\arraybackslash}m{1.80cm}>{\centering\arraybackslash}m{1.80cm}}
    \specialrule{.1em}{.05em}{0.0em}
         &  \multicolumn{3}{c|}{Contact estimation} & \multicolumn{4}{c}{3D reconstruction} \\ 
         Contact estimation methods & Contact$_{\text{p}}{\uparrow}$ & Contact$_{\text{r}}{\uparrow}$ & Contact$_{\text{f1}}{\uparrow}$ & CD$_{\text{human}}{\downarrow}$ & CD$_{\text{object}}{\downarrow}$ & Contact$_{\uparrow}$ & Collision${\downarrow}$ \\ \hline
        w/o contact & -- & -- & -- & 5.311 & 19.849 & 0.374 & 0.054 \\ \cdashline{1-8}
        $P_{\text{contact}}$ & 0.282 & 0.342 & 0.309 & 4.941 & 16.701 & 0.412 & 0.047 \\
        DECO~\cite{tripathi2023deco} & 0.200 & 0.264 & 0.228 & 5.115 & 17.229 & 0.353 & 0.051 \\
        LEMON~\cite{yang2024lemon} & 0.426 & 0.225 & 0.295 & 5.084 & 17.060 & 0.389 & 0.050  \\
        InteractVLM~\cite{dwivedi2025interactvlm} & 0.422 & 0.458 & 0.439 & 4.988 & 16.009 & 0.408 & 0.052\\
        \specialrule{.1em}{-0.05em}{-0.05em}
    \end{tabular}
    \vspace*{-1mm}
}
\end{center}
    \vspace*{-1.3em}
    \caption{
    \textbf{Impact of contact estimation accuracy on \ourmethod's reconstruction, evaluated on Open3DHOI~\cite{wen2025reconstructing}.}
    }
    \vspace*{-0.6em}
    \label{tab:suppl_contact_impact}
\end{table*}
\section{Comparison with HDM}
\label{sec:suppl_hdm}
\cref{tab:suppl_hdm} shows that our framework significantly outperforms HDM~\cite{xie2023templatefree}, a state-of-the-art method that reconstructs 3D human and object shapes as point clouds rather than meshes.
Since HDM does not estimate SMPL-X mesh surfaces, the standard root-based alignment described in \cref{sec:evaluation_metrics} cannot be directly applied for the evaluation. 
Instead, we employ the Iterative Closest Point (ICP) algorithm to align the reconstructed human vertices with the ground-truth (GT) vertices.
Consequently, HDM, which is a learning-based approach, is particularly vulnerable to unseen object categories compared to those encountered during training (\textit{e.g.}, balls and suitcases).
On the other hand, since our framework is an optimization-based approach, it generalizes effectively to unseen objects by leveraging the strong prior knowledge of the diffusion network.
\begin{figure}[t!]
  \vspace*{-0.1em}
  \centering
  \includegraphics[width=0.78\linewidth]{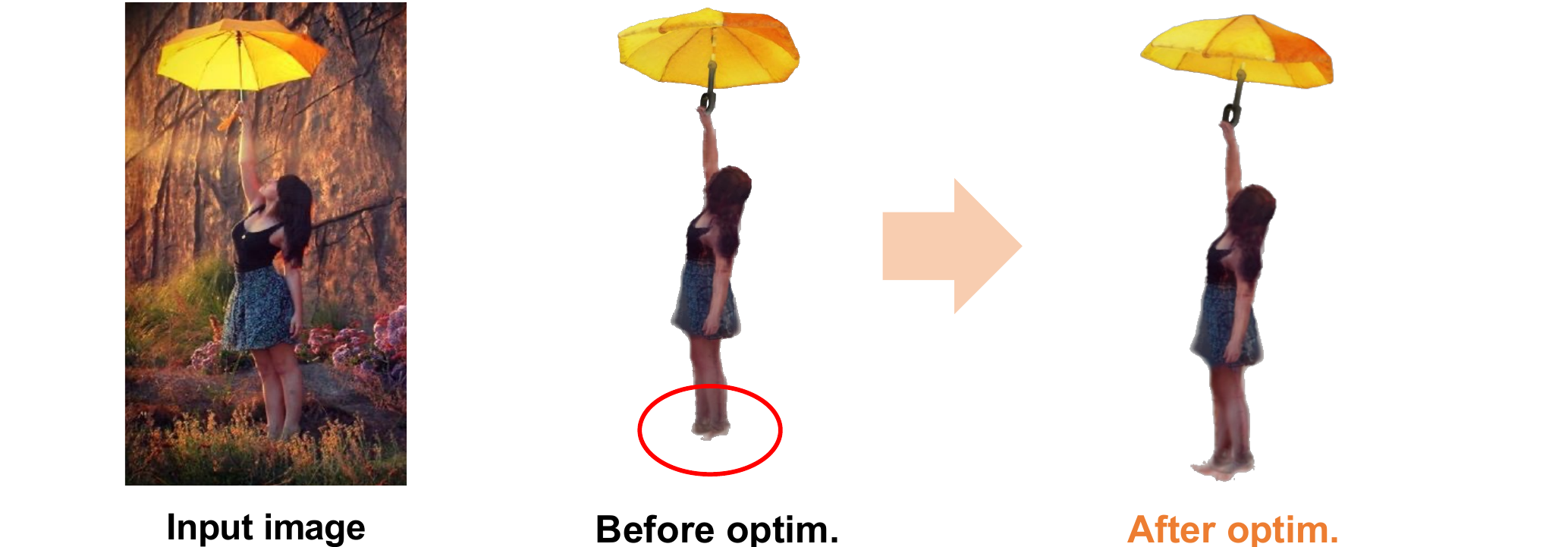}
  \vspace*{-0.1em}
  \caption{\textbf{Enhancement of Gaussian via optimization process.}
  }
  \vspace*{+0.0em}
  \label{fig:suppl_gaussian_enhance}
\end{figure}

\section{Impact of contact estimation accuracy}
\label{sec:suppl_contact}
\cref{tab:suppl_contact_impact} illustrates the impact of contact estimation accuracy on the 3D reconstruction performance of our framework.
We compare our framework under different contact estimation settings, including specialized contact prediction models such as LEMON~\cite{yang2024lemon} and InteractVLM~\cite{dwivedi2025interactvlm}.
While these models improve the precision of contact localization on human and object surfaces, their contribution to final 3D reconstruction quality remains marginal.
The contact estimation methods primarily focus on accurately predicting the boundaries of the contact region at a fine-grained level.
However, regardless of how precise these contact boundaries are, the contact information alone cannot capture the holistic and semantic context of human-object interaction.
\begin{table}[t]
\def\arraystretch{1.55}
\renewcommand{\tabcolsep}{0.8mm}
\footnotesize
\begin{center}
\scalebox{0.78}{
    \begin{tabular}{>{\raggedright\arraybackslash}m{3.25cm}|>{\centering\arraybackslash}m{1.35cm}>{\centering\arraybackslash}m{1.35cm}>{\centering\arraybackslash}m{1.35cm}>{\centering\arraybackslash}m{1.35cm}}
    \specialrule{.1em}{.05em}{0.0em}
        &  \multicolumn{4}{c}{Open3DHOI} \\
         & CD$_{\text{human}}{\downarrow}$ & CD$_{\text{object}}{\downarrow}$  & Contact${\uparrow}$ & Collision${\downarrow}$  \\ \hline
        Before conversion & 5.020 & 16.987 & 0.394 & 0.052  \\
        \textbf{After conversion~(Ours)} & \textbf{4.941} & \textbf{16.701} & \textbf{0.412} & \textbf{0.047} \\
        \specialrule{.1em}{-0.05em}{-0.05em}
    \end{tabular}
    \vspace*{-1mm}
}
\end{center}
    \vspace*{-1.2em}
    \caption{
    \textbf{Impact of Gaussians-to-mesh conversion}
    }
    \vspace*{-0.3em}
    \label{tab:suppl_gaussian-to-mesh}
\end{table}
This observation suggests that capturing the holistic interaction context is far more important for the joint reconstruction of 3D human and object than precisely delineating fine-grained contact boundaries.
Accordingly, the core strength of our framework is determined by holistic contact reasoning supported by text-guided optimization rather than by accurate contact prediction.
This validates our use of the contact text prompt $P_{\text{contact}}$ as a lightweight yet effective alternative to external contact prediction models.

\section{Impact of Gaussian attributes optimization}
\label{sec:suppl_gs_optim}
\cref{fig:suppl_gaussian_enhance} demonstrates the importance of optimizing the 3D human and object Gaussian attributes~($\phi_{\text{h}}$ and $\phi_{\text{o}}$) within our framework.
Since the initial Gaussian attributes can occasionally be incomplete due to occlusions in the input image, we further refine them through the appearance loss $L_{\text{appr}}$.
This optimization process enhances the visual plausibility and overall coherence of the reconstructed human–object interactions.
As there is no existing 3D HOI dataset that provides both geometry and texture annotations, quantitative evaluation of this optimization remains challenging; thus, we primarily present qualitative results.

\section{Impact of Gaussians-to-mesh conversion}
\label{sec:suppl_g2m}
\cref{tab:suppl_gaussian-to-mesh} shows that our Gaussians-to-mesh conversion process, detailed in Sec.~\textcolor{red}{3.4}, is a crucial step for accurate mesh reconstruction.
Direct conversion of 3D Gaussians to mesh surfaces often produces inconsistencies near contact regions. Accordingly, we use the conversion procedure that enforces geometric consistency between Gaussian-defined contact regions and the corresponding mesh vertices.
As a result, it improves overall geometric accuracy and yields substantial gains in the contact evaluation score ($\text{Contact}$).

\section{Details of text captioning}
\label{sec:suppl_captioning}
\cref{fig:suppl_captioning_1} and \cref{fig:suppl_captioning_2} illustrate the two captioning instructions used to generate the holistic text prompt $P_{\text{holistic}}$ and contact text prompt $P_{\text{contact}}$, with the GPT-4~\cite{achiam2023gpt} vision-language model (VLM).
First, we generate the holistic prompt $P_{\text{holistic}}$, which describes the interaction between the person closest to the image center and the object most directly involved with that person.
Then, we generate the contact prompt $P_{\text{contact}}$ by providing both the input image and the holistic description $P_{\text{holistic}}$ as inference cues, enabling it to infer which human body parts are in direct physical contact with the object.
This two-stage captioning strategy encourages each stage to specialize in a distinct role: inferring global interaction semantics and localized contact information, respectively.

\begin{figure*}[t!]
  \centering
  \includegraphics[width=1.0\linewidth]{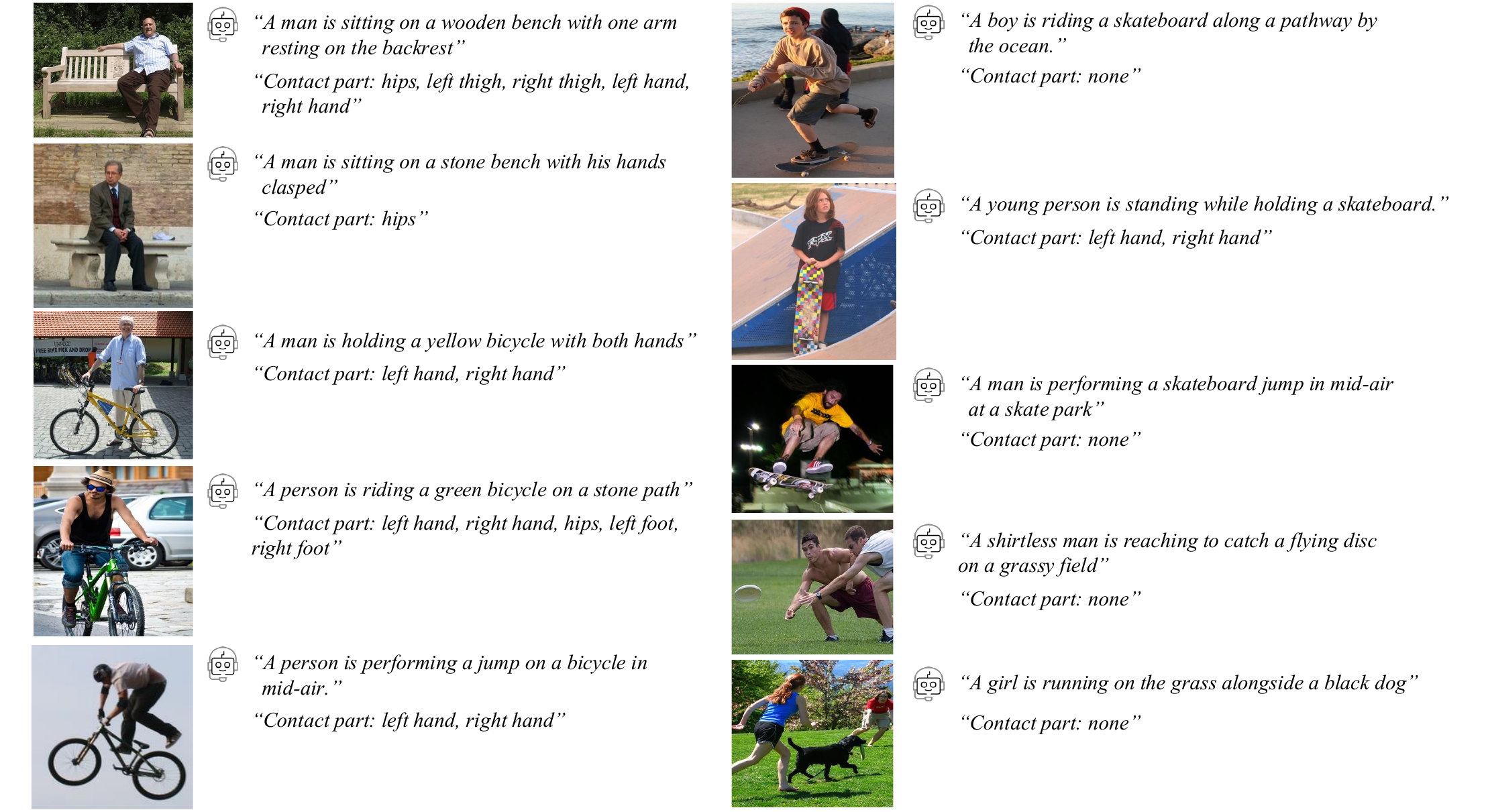}
  \vspace*{-1.4em}
  \caption{\textbf{Text captioning results on Open3DHOI~\cite{wen2025reconstructing}.}
  Our text captioning produces accurate and rich text descriptions for a wide range of interaction scenarios.
  }
  \vspace*{-0.5em}
  \label{fig:suppl_captioning_results}
\end{figure*}

\cref{fig:suppl_captioning_results} highlights the strong capability of the text captioning process.
As shown in the examples, it successfully captures key contextual cues essential for reasoning about human–object interactions, including the human’s action (\textit{e.g.}, sitting, riding, and performing) and the surrounding environment (\textit{e.g.}, pathway, mid-air, and grassy field).
Even when the same object appears in different interaction scenarios, the VLM provides accurate and semantically appropriate descriptions.
This demonstrates the richness of holistic contextual information, in contrast to contact cues that convey only local geometric proximity.
Such comprehensive interaction cues play a crucial role in guiding our reconstruction framework toward more accurate and globally coherent 3D human and object reconstructions.

\begin{figure}[t!]
    \begin{lstlisting}[]
### TASK ###
Your goal is to provide a detailed description of the given image, which depicts the interaction between a person and an object.

- Focus only on the person whose body center is closest to the image center.
- Identify the object most directly interacted with and state the action.
- Output must be one sentence, no explanations, labels, or reasoning.
- Additionally, explicitly output the single object that is most directly interacted with.

### OUTPUT FORMAT ###
Output: {{interacting object}}, {{description}}

### OUTPUT EXAMPLE ###
Example 1 - Output: soccer ball, A woman is playing soccer on a grassy field, dribbling the ball.
Example 2 - Output: small box, A man is seated on a small box with legs crossed.
Example 3 - Output: chair, A woman is moving a chair with one hand.
    \end{lstlisting}
    \vspace{-0.8em} 
\caption{\textbf{Captioning instruction for the VLM~\cite{achiam2023gpt} to acquire holistic text prompt $P_{\text{holistic}}$.}
}
\vspace*{-0.2em}
\label{fig:suppl_captioning_1}
\end{figure}

\section{Implementation details}
\label{sec:suppl_implementation}
We explain the implementation details of two stages: the reconstruction stage (Sec.~\textcolor{red}{3.2}) and HOI optimization stage (Sec.~\textcolor{red}{3.3}), below.
PyTorch~\cite{paszke2017automatic} is used for implementation.

\begin{figure}[t]
    \begin{lstlisting}[]
### TASK ###
Your goal is to list the body parts of the person in the given image that are in direct physical contact with the object.

- Choose ONLY from this pre-defined list   (multi-select allowed): head, hips, ...
- The interacting object and reference description are provided as follows: "(*@\textbf{\{object\}}@*)", "(*@\textbf{\{description\}}@*)".
- Focus only on the person whose body center is closest to the image center.
- Identify the object most directly interacted with and state the action.
- LEFT/RIGHT must be relative to the person (egocentric), not the viewer/camera.
- If no clear physical contact is visible, output none.

### OUTPUT FORMAT ###
Output: {{comma-separated list}}

### OUTPUT EXAMPLE ###
Example 1 - Output: left hand, right hand
Example 2 - Output: right foot
Example 3 - Output: none
    \end{lstlisting}
    \vspace{-0.8em} 
\caption{\textbf{Captioning instruction for the VLM~\cite{achiam2023gpt} to acquire contact text prompt $P_{\text{contact}}$.}
}
\vspace*{-0.2em}
\label{fig:suppl_captioning_2}
\end{figure}

\subsection{Reconstruction stage}
\noindent\textbf{Human reconstruction.}
When using SmartEraser~\cite{jiang2025smarteraser}, the object regions to be removed are inpainted using classifier-free guidance with a guidance scale of 1.5 in its generative diffusion network.
To segment human region from the inpainted image, we use Grounded-SAM~\cite{liu2023grounding,kirillov2023segment} with a text prompt corresponding to the object category name obtained from the text captioning~(\cref{sec:suppl_captioning}).
From the segmented human image, LHM operates on a canonical set of 40,000 Gaussian anchors uniformly sampled over the SMPL-X surface.
For each anchor, LHM predicts the Gaussian attributes $\phi_{\text{h}}$, including canonical offsets, opacity, scale, and appearance features, through a single feed-forward inference.

\begin{figure}[t!]
  \centering
  \includegraphics[width=0.9\linewidth]{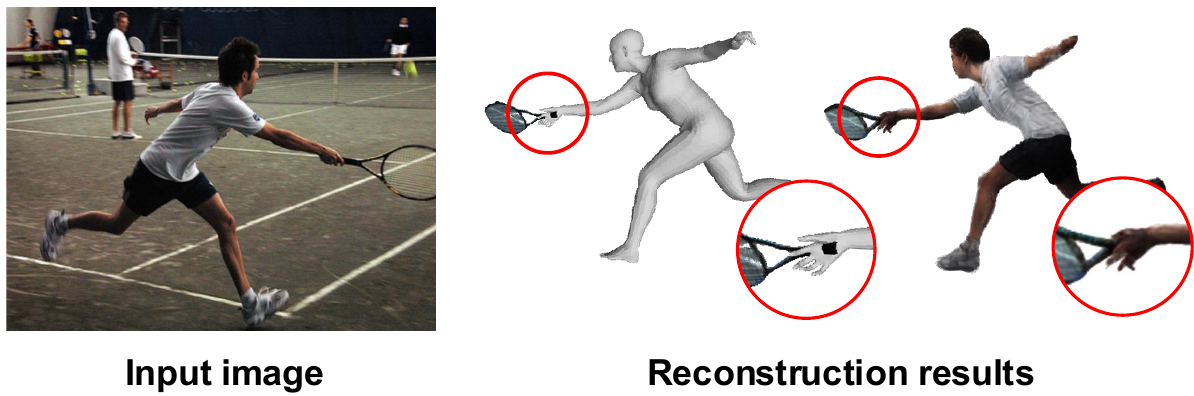}
  \vspace*{-0.34em}
  \caption{\textbf{Failure case of reconstructing local details.}
  }
  \vspace*{-0.4em}
  \label{fig:suppl_limitations}
\end{figure}

\noindent\textbf{Object reconstruction.}
When using SmartEraser~\cite{jiang2025smarteraser}, we adopt the same settings as in the human reconstruction stage.
To segment object region from the inpainted image, we use Grounded-SAM~\cite{liu2023grounding,kirillov2023segment} with the text prompt “human”.
From the segmented object image, InstantMesh~\cite{xu2024instantmesh} first synthesizes six multi-view images using Zero123++~\cite{shi2023zero123plus} with 75 diffusion steps, and then reconstructs a textured mesh through its triplane-based reconstruction network.
The resulting textured mesh is subsequently converted into 3D object Gaussians $\phi_{\text{o}}$, where the Gaussian centroids are placed at the mesh vertex positions and their initial appearance features are assigned from the mesh vertex colors.
The Gaussian attributes are further optimized to match the 2D images rendered from the reconstructed textured mesh at 360 uniformly sampled viewpoints, following the optimization procedure of 3DGS~\cite{kerbl3Dgaussians}.

\subsection{HOI optimization stage}
We use the Adam~\cite{kingma2015adam} optimizer with an exponentially decaying learning rate.
The initial learning rate is set to $1 \times 10^{-2}$ for the object pose parameters ($R$, $t$, and $s$), $1 \times 10^{-4}$ for the human pose parameters ($\theta$ and $\beta$), and $1 \times 10^{-4}$ for the human and object Gaussian attributes ($\phi_{\text{h}}$ and $\phi_{\text{o}}$).
We run the optimization for $N=200$ steps on a single NVIDIA RTX 8000 GPU.
Under this setting, the average optimization time per sample is 134 seconds.

\noindent\textbf{Appearance rendering.}
During optimization, we render the 3D human Gaussians $\Phi_{\text{h}}$ and object Gaussians $\Phi_{\text{o}}$ using a spherical coordinate system $(r, \upsilon, \psi)$, where $r$ denotes the distance to the spherical origin, $\upsilon$ the elevation angle, and $\psi$ the azimuth angle.
We uniformly sample viewpoints with $r \in [1.0, 2.5]$, $\upsilon \in [-30^{\circ}, 30^{\circ}]$, and $\psi \in [-180^{\circ}, 180^{\circ}]$.
Since human-object interaction primarily involves the upper body, such as the head and hands, we additionally use zoomed-in camera views focused on this region. For these upper-body views, we set the spherical origin to the 3D position of the SMPL-X spine keypoint and sample $r \in [0.7, 1.5]$, $\upsilon \in [-30^{\circ}, 30^{\circ}]$, and $\psi \in [-180^{\circ}, 180^{\circ}]$.

\noindent\textbf{Appearance loss.}
We compute the appearance loss $\mathcal{L}_{\text{appr}}$ of Eq.~(\textcolor{red}{2}) using StableDiffusion-v2.1~\cite{rombach2022high} and apply classifier-free guidance~\cite{ho2021cfg} with a guidance scale of $15.0$ for noise estimation.
The noise levels are defined at randomly sampled timesteps within $[0.02, 0.98]$.
To ensure stable optimization, we clip loss gradients to a maximum norm of $1.0$.

\section{Limitations and future work}
\label{sec:suppl_limitation}

\noindent\textbf{Reconstruction of local details.}
While our framework captures holistic human–object interactions effectively, it may overlook fine-grained local details such as small accessories or subtle surface deformations, as shown in \cref{fig:suppl_limitations}.
This limitation occurs because the appearance loss of our framework primarily offers global guidance and lacks fine-grained supervision that specifically addresses local regions.
A promising future direction is to design localized, text-driven supervision that specializes in local regions to further enhance fine-detail reconstruction.

\noindent\textbf{Video as input.}
Our framework aims to jointly reconstruct 3D human and object from a single image.
When extending the method to video input, additional considerations become essential, such as maintaining temporal consistency across frames and ensuring consistent geometry and texture over time.
With the recent emergence of text-to-video generative models~\cite{ho2022video,blattmann2023align}, future work could leverage these advances by using text descriptions as a key guidance, enabling more stable and temporally coherent 3D HOI reconstruction.

\section{More qualitative results}
\label{sec:suppl_more_results}
\begin{figure*}[t]
  \centering
  \includegraphics[width=0.95\linewidth]{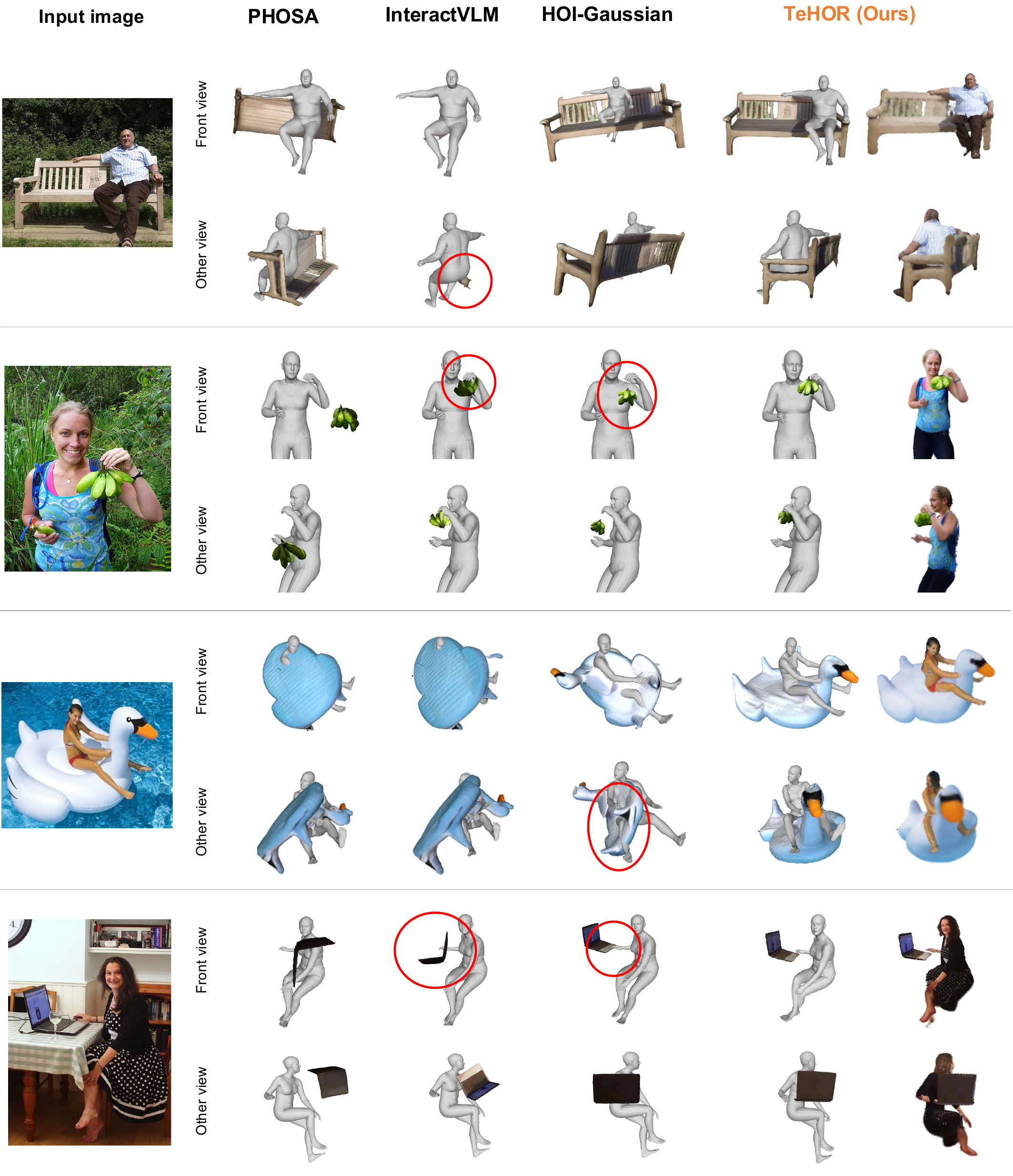}
  \vspace{-0.5em}
  \caption{\textbf{
  More qualitative comparison of 3D human and object reconstruction with PHOSA~\cite{zhang2020perceiving}, InteractVLM~\cite{dwivedi2025interactvlm}, and HOI-Gaussian~\cite{wen2025reconstructing}, on Open3DHOI~\cite{wen2025reconstructing}.}
  We highlight their representative failure cases with red circles. 
  }
  \label{fig:more_results_1}
\end{figure*}

\begin{figure*}[t]
  \centering
  \includegraphics[width=0.95\linewidth]{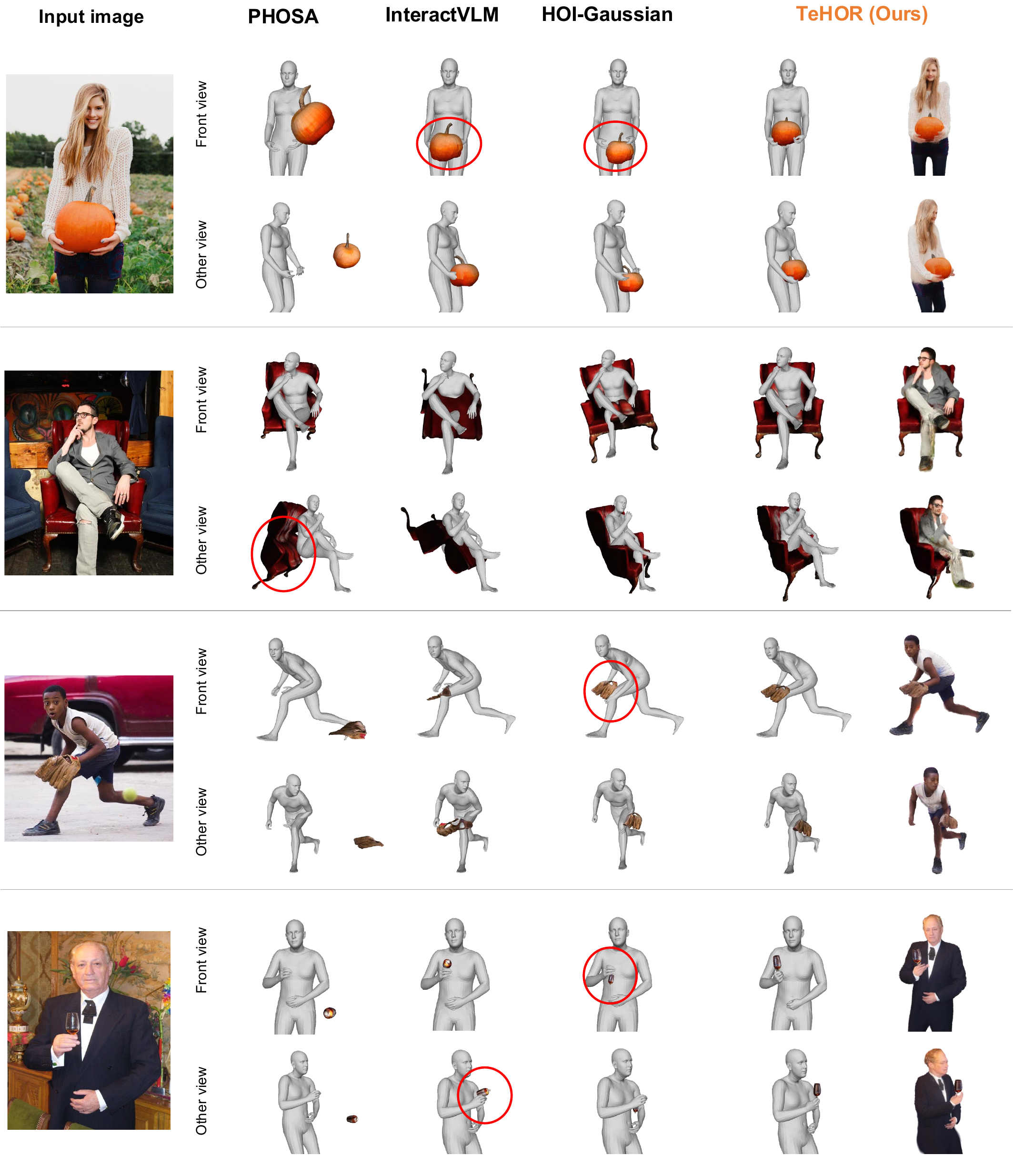}
  \vspace{-0.5em}
  \caption{\textbf{
  More qualitative comparison of 3D human and object reconstruction with PHOSA~\cite{zhang2020perceiving}, InteractVLM~\cite{dwivedi2025interactvlm}, and HOI-Gaussian~\cite{wen2025reconstructing}, on Open3DHOI~\cite{wen2025reconstructing}.}
  We highlight their representative failure cases with red circles. 
  }
  \label{fig:more_results_2}
\end{figure*}

\begin{figure*}[t]
  \centering
  \includegraphics[width=0.95\linewidth]{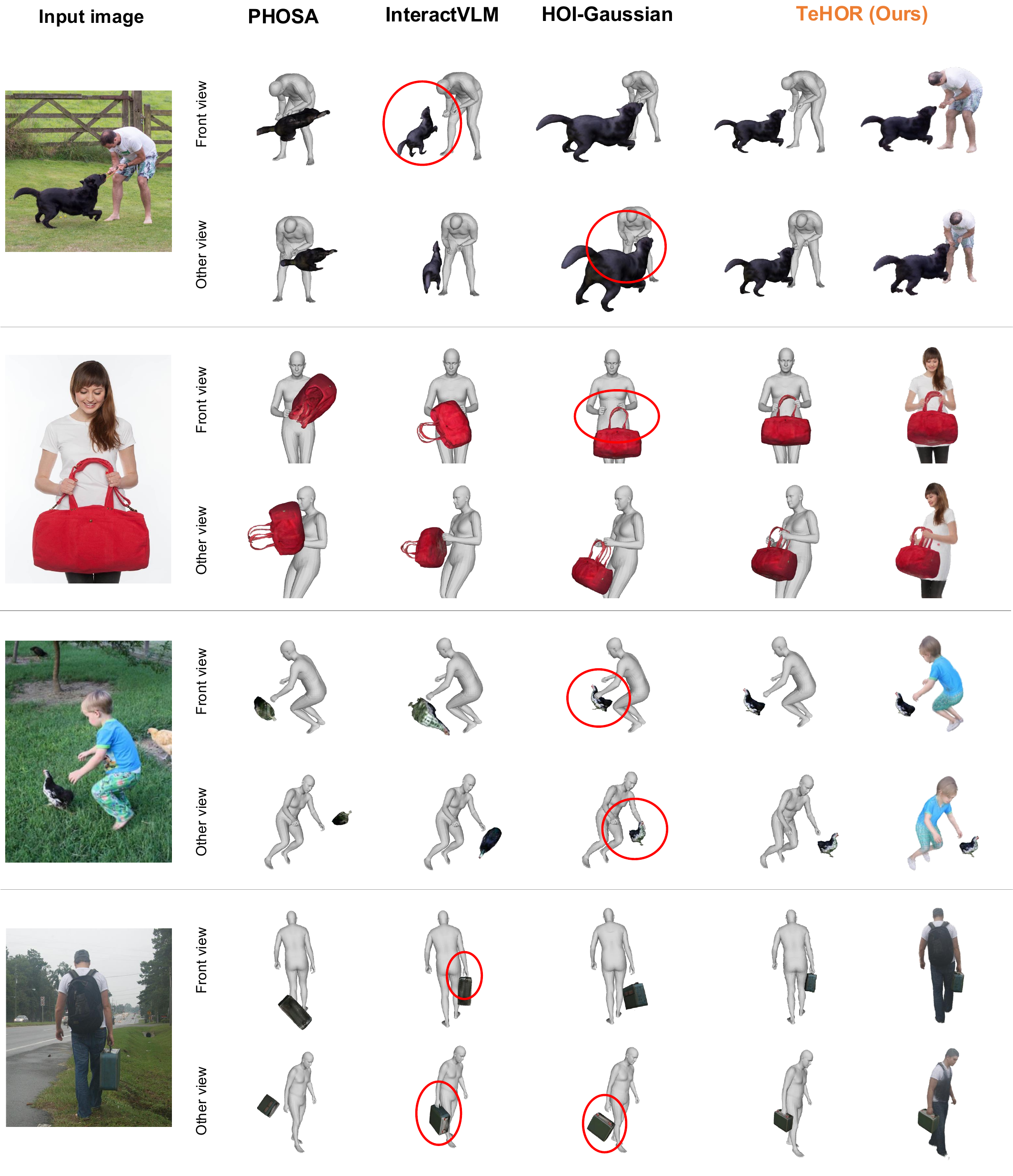}
  \vspace{-0.5em}
  \caption{\textbf{
  More qualitative comparison of 3D human and object reconstruction with PHOSA~\cite{zhang2020perceiving}, InteractVLM~\cite{dwivedi2025interactvlm}, and HOI-Gaussian~\cite{wen2025reconstructing}, on Open3DHOI~\cite{wen2025reconstructing}.}
  We highlight their representative failure cases with red circles. 
  }
  \label{fig:more_results_3}
\end{figure*}

\begin{figure*}[t]
  \centering
  \includegraphics[width=0.95\linewidth]{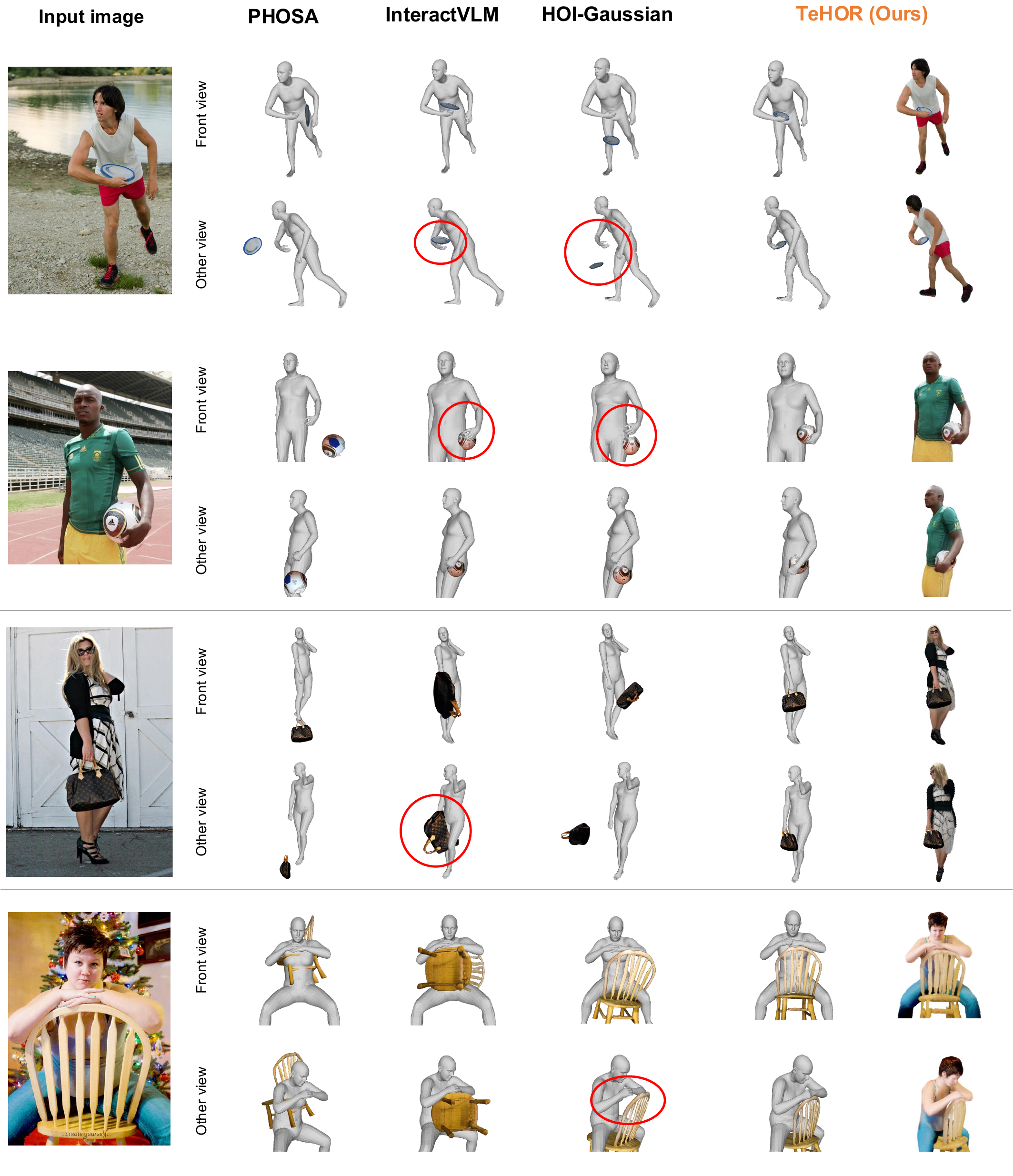}
  \vspace{-0.5em}
  \caption{\textbf{
  More qualitative comparison of 3D human and object reconstruction with PHOSA~\cite{zhang2020perceiving}, InteractVLM~\cite{dwivedi2025interactvlm}, and HOI-Gaussian~\cite{wen2025reconstructing}, on Open3DHOI~\cite{wen2025reconstructing}.}
  We highlight their representative failure cases with red circles. 
  }
  \label{fig:more_results_4}
\end{figure*}

We provide additional qualitative comparison results of our \ourmethod~in~\cref{fig:more_results_1,fig:more_results_2,fig:more_results_3,fig:more_results_4}.
These examples further demonstrate the effectiveness of our method in reconstructing realistic and semantically coherent human–object interactions.
Please note that the left-side results of \ourmethod~are mesh-based renderings, while the right-side results are Gaussian-based renderings.
Due to the inherent characteristics of 3D Gaussian representations, Gaussian renderings can appear slightly larger and exhibit blurred boundaries.

\clearpage

\clearpage

\clearpage

\clearpage

\clearpage
{
    \small
    \bibliographystyle{ieeenat_fullname}
    \bibliography{egbib}
}
\end{document}